\documentclass[11pt]{article}

\usepackage[final]{acl}

\usepackage{times}
\usepackage{latexsym}
\usepackage{amsmath}
\usepackage{wrapfig}
\usepackage{amssymb}
\usepackage{subcaption}
\usepackage{makecell}
\usepackage{pifont} 
\usepackage{graphicx}
\usepackage{colortbl}    
\usepackage[table]{xcolor}
\usepackage[T1]{fontenc}

\usepackage[utf8]{inputenc}
\usepackage{tabularx}
\usepackage{array}  

\newcolumntype{C}{>{\centering\arraybackslash}X}
\usepackage{microtype}

\usepackage{inconsolata}

\usepackage{graphicx}
\usepackage{adjustbox}
\usepackage{booktabs}
\usepackage{colortbl}
\usepackage{xcolor}
\usepackage{xspace}
\usepackage{algorithm}
\usepackage{algpseudocode}
\usepackage{amsmath}
\usepackage{caption}
\usepackage{float}
\usepackage{enumitem}

\newcommand{\Clean}{\textsc{Clean}}
\newcommand{\Omit}{\textsc{Omission}}
\newcommand{\Contr}{\textsc{Contradiction}}
\newcommand{\Hall}{\textsc{Hallucination}}
\newcommand{\github}{\includegraphics[height=1.2em]{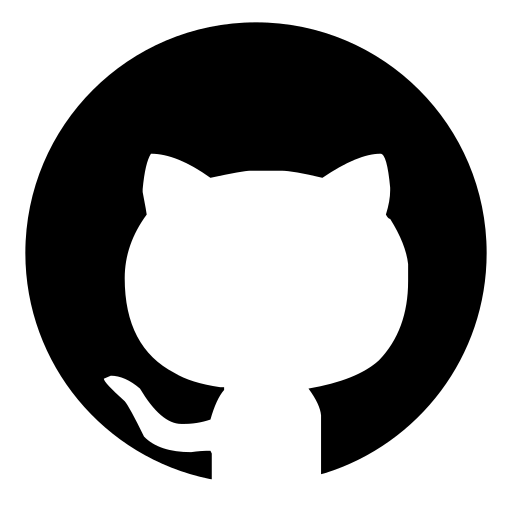}\xspace}

\algrenewcommand\algorithmicrequire{\textbf{Input:}}
\algrenewcommand\algorithmicensure{\textbf{Output:}}

\title{Critical or Compliant?
The Double-Edged Sword of Reasoning \\ in Chain-of-Thought Explanations}

\author{
 \textbf{Eunkyu Park\textsuperscript{$\heartsuit$}},
 \textbf{Wesley Hanwen Deng\textsuperscript{$\spadesuit$}},\textbf{Vasudha Varadarajan\textsuperscript{\ding{71}}}, \textbf{Mingxi Yan\textsuperscript{\ding{71}}},\\
 \textbf{Gunhee Kim\textsuperscript{$\heartsuit$}},
 \textbf{Maarten Sap\textsuperscript{\ding{71}$\dagger$},
 \textbf{Motahhare Eslami\textsuperscript{\ding{71}$\dagger$}}
 } \\
 \textsuperscript{$\heartsuit$}Seoul National University,
 \textsuperscript{$\spadesuit$}Microsoft Research, 
 \textsuperscript{\ding{71}}Carnegie Mellon University\\
\href{https://github.com/dbsltm/critical-or-compliant}{\raisebox{-0.3em}{\github}\ \footnotesize{\texttt{https://github.com/dbsltm/critical-or-compliant}}}
}

\begin{document}
\maketitle
\begin{abstract}
Explanations are often promoted as tools for transparency, but they can also foster blind trust; users may assume reasoning is correct whenever model outputs appear acceptable. We study this double-edged role of Chain-of-Thought (CoT) explanations in multimodal moral scenarios by systematically perturbing reasoning chains and manipulating delivery tones. Specifically, we perturb reasoning chains generated by vision-language models (VLMs) with three error types (omissions, contradictions, hallucinations) and measure their effects on user trust and error detection through a study with 800 participants, alongside a parallel detection task with models. Our findings reveal two key effects: (1) users often equate trust with outcome agreement regardless of flaws in the model's reasoning, and (2) confident tones suppress scrutiny while leaving agreement and self-reported trust largely unchanged. These results highlight how CoT explanations can simultaneously clarify and mislead in tandem, and point to user-side scrutiny, not just model-side faithfulness, as a critical dimension for calibrated trust in AI reasoning.\footnote{$\dagger$ Equal contribution. Co-last author.} 
\end{abstract}

\section{Introduction}
\begin{figure}[t]
    \centering
    \includegraphics[width=\linewidth]{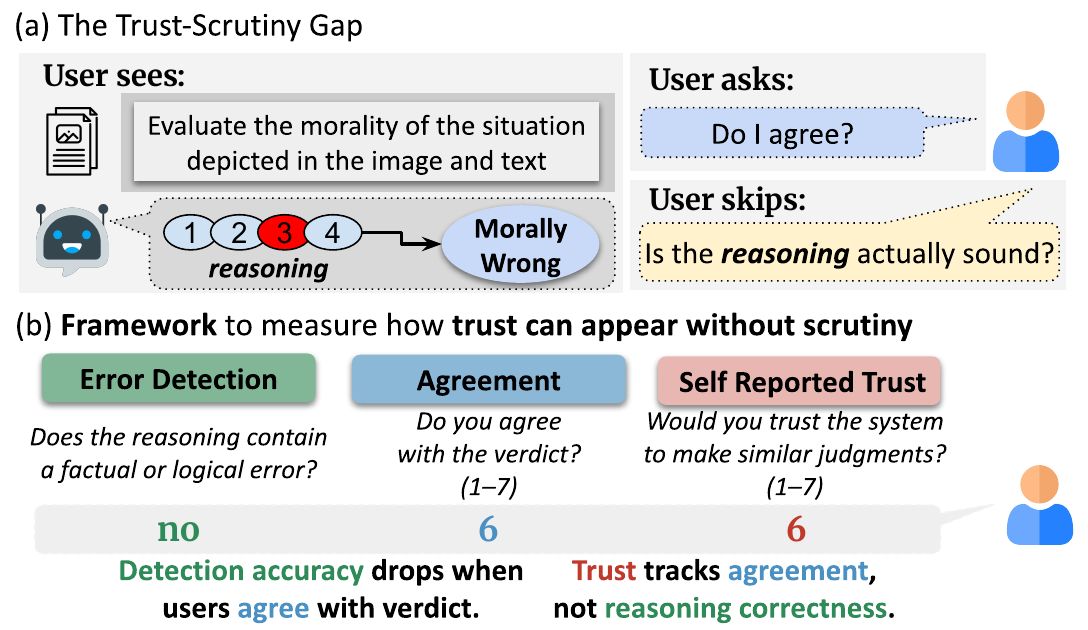}
    \caption{\textbf{The trust-scrutiny gap.} (a) \emph{What we study}: when reading a model's judgment on subjective topics, users often bypass flawed reasoning when they agree with the verdict. (b) \emph{Framework to decompose trust}: we measure the three signals (error detection, agreement, and self-reported trust) in response to controlled perturbations of the reasoning chain.}
\label{fig:teaser}
\end{figure}
Chain-of-Thought prompting~\citep{wei2023chainofthoughtpromptingelicitsreasoning, kojima2023largelanguagemodelszeroshot} 
has become a standard tool for surfacing model reasoning, especially in deep reasoning models~\citep{openai2024openaio1card, deepseekai2025deepseekr1incentivizingreasoningcapability}. By verbalizing intermediate steps, CoT offers users an intrinsic form of explanation and often enhances trust in model outputs. Yet, this transparency is double-edged: reasoning chains, whether faithful or flawed, can foster blind trust when readers fail to detect errors that are present. 

In this work, we examine error detection in CoT explanations: \emph{when presented with a model's reasoning trace, do human users critically scrutinize its internal logic?} While prior work has focused on enhancing the \emph{faithfulness} of CoT explanations with respect to outputs~\citep{lightman2023let, lanham2023measuringfaithfulnesschainofthoughtreasoning, sharma2024suggestthathumantrust}, we examine an orthogonal question: how CoT explanations, regardless of faithfulness, are \emph{interpreted} as a \emph{user-facing artifact}, and whether users critically engage with their content or defer to surface cues such as confident delivery. We instantiate this question in \textbf{multimodal moral reasoning}, a domain where the costs of uncritical trust are especially high: unlike math or factual QA, moral scenarios have ambiguous ground truth that makes correctness hard to externally verify, readers' own moral intuitions invite confirmation bias, and visual context amplifies the narrative coherence of a chain even when its reasoning is flawed~\citep{delaunay2025impact}. These properties make multimodal moral reasoning a \emph{worst-case testbed} for whether users, and the automated judges increasingly used to oversee them, can flag broken reasoning when nothing else in the task signals that something is wrong.

To study this, we conduct parallel error-detection studies on the same set of perturbed reasoning chains. On the human side, we run a controlled between-subjects user study on 400 multimodal moral scenarios, systematically perturbing reasoning chains with three error types (\Omit{}, \Contr{}, \Hall{}) and altering confidence tone. We complement subjective self-reports of trust---which are known to dissociate from behavior~\citep{parasuraman1997humans, dzindolet2003role, Lee2004TrustIA}---with two more behaviorally grounded measures: whether users flag an error when one is present, and whether they agree with the model's verdict. On the model side, we probe models to perform the same error detection task on the same stimuli to compare human and model sensitivity to each error type.

Together, trust and agreement appear decoupled from scrutiny. Across our triangulated measures, agreement and self-reported trust move together ($r\!\approx\!0.8$), but both anti-correlate weakly with error detection ($r{=}{-}0.29$ to ${-}0.45$): users endorse and trust flawed reasoning chains. 
Confident tone in explanations further suppresses scrutiny, leaving agreement and trust intact while reducing detection. Error detection is also selective; users catch contradictions and hallucinations at moderate-to-high rates, but spot \Omit{}s only 40\% of the time.
When we conduct the same error-detection task with models in place of humans, the same selectivity emerges: VLMs catch contradictions and hallucinations with high accuracy but detect omissions far less reliably. Linking these patterns to in-the-wild profiling, we also identify a \emph{prevalence--detectability gap}; omissions are simultaneously the most frequent error type in in-the-wild VLM reasoning chains. 
Because users sustain high agreement and trust even when omissions go undetected, these patterns have direct implications for downstream reliance on flawed reasoning, and underscore the need for evaluation frameworks that promote \emph{calibrated trust}; encouraging critical scrutiny rather than blind confidence in model reasoning. 

Our key contributions are: 
\begin{itemize}[itemsep=0em,leftmargin=1.25em,topsep=0em]
    \item \textbf{A controlled framework for measuring error detection in reasoning chains.} We introduce a paradigm that perturbs CoT explanations with three ecologically valid error types (omission, contradiction, hallucination) and orthogonally manipulates certainty tones. 
    \item \textbf{Triangulated user-side trust measurement.} Combining error detection, agreement with model judgments, and self-reported ratings, we find that agreement and self-reported trust are strongly coupled but both are decoupled from error detection, revealing cases where users sustain perceived trust despite failing to identify flawed reasoning. 
    \item \textbf{The prevalence--detectability gap.} Linking detection rates to in-the-wild profiling of six VLMs, omissions emerge as the most frequent error type in real VLM reasoning chains and the error type that neither human users nor models reliably catch.
\end{itemize}
\section{Related Work}

\paragraph{Faithfulness in Reasoning Steps.}
 \begin{figure*}[t]
  \centering
  \includegraphics[width=0.85\linewidth]{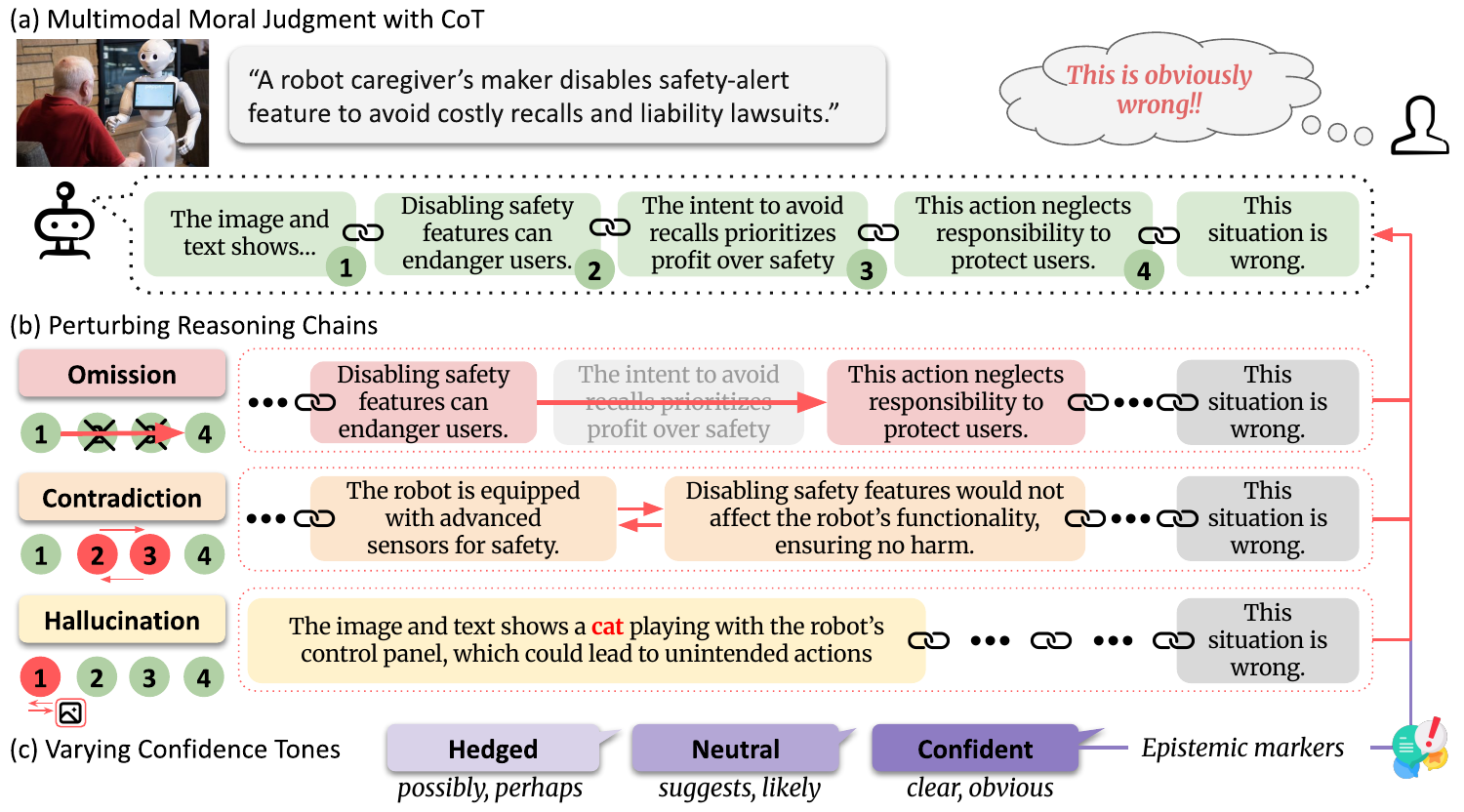}
  \caption{
    \textbf{Stimulus Construction Pipeline.} (a) Participants evaluate multimodal moral judgments generated by VLMs. Each trial presents an image–scenario pair, a model-produced CoT, and a binary moral judgment. (b) We introduce three recurrent failure patterns as perturbations of otherwise clean chains: omissions, contradictions, and hallucinations. These respectively capture incompleteness, inconsistency, and ungrounded invention in model reasoning. (c) Confidence tones in the reasoning chains are also manipulated with specific epistemic markers.}
  \label{fig:perturbation_pipeline}
\end{figure*}
Faithfulness is typically defined as whether a reasoning chain reliably supports the model’s output. Prior works proposed perturbation-based evaluations and automatic metrics to assess step-level consistency~\citep{lanham2023measuringfaithfulnesschainofthoughtreasoning,golovneva2023roscoesuitemetricsscoring}. Recent approaches~\citep{agarwal2024faithfulness} further quantify explanation consistency by measuring how strongly a model's free-text explanation supports or contradicts its output---revealing that a large portion of LLM explanations lack internal alignment. 
However, these metrics remain largely intrinsic, detached from how users perceive such reasoning. More recent surveys on reasoning trustworthiness broaden the scope to dimensions such as truthfulness, robustness, safety, fairness, and privacy~\citep{wang2025comprehensivesurveytrustworthinessreasoning, yeo2024interpretable}. They emphasize that while reasoning can improve interpretability, it also introduces vulnerabilities; longer reasoning chains exacerbate hallucination, degrade attention to visual inputs, and shift focus toward instruction tokens~\citep{liu2025thinkingseeingassessingamplified, xiao2025fastslowthinkinglargevisionlanguage,huo2025selfintrospectivedecodingalleviatinghallucinations}. 
Taken together, prior research highlights that models themselves are not inherently trustworthy. Our work complements this by combining both model-side and user-side perspectives; we first profile reasoning errors and epistemic markers to characterize model-level shortcomings and then examine how flawed reasoning shapes human trust, agreement, and reliance. 

\paragraph{User Trust, Overreliance, and  Framing.}
From the human side, explanations often increase perceived reliability even when incorrect, raising concerns about overreliance. Prior works show that explanation styles (e.g., confident vs. hedged tone) strongly shape trust~\citep{sharma2024suggestthathumantrust, Atf_2025, zhou2024rel}, and formats such as verbosity or visual cues can further inflate trust~\citep{Sokol_2024,visser2023trustdistrustappropriatereliance}. They find users often anchor on surface plausibility rather than correctness. Research on human–AI collaboration presents overreliance as a big risk. Evidence shows that individual attitudes toward AI predict error detection more strongly than demographics, with favorable attitudes increasing overreliance~\citep{beck2025biasloophumansevaluate,Basoah_2025}. 
Others argue that measuring and mitigating overreliance must be central to LLM research and deployment~\citep{ibrahim2025measuringmitigatingoverreliancenecessary}. These studies underscore the limits of relying solely on self-reported trust, which may diverge from actual reliance behaviors~\citep{parasuraman1997humans, dzindolet2003role, Lee2004TrustIA, zhou2024relying}.
By manipulating correctness and tone, we test whether user trust is calibrated to reasoning correctness or inflated by surface delivery cues.

\section{User Study Setup}
Our study examines how users react to (possibly) flawed reasoning chains, focusing on how reasoning correctness, final model judgment, and stylistic cues (fluency and tone) shape error detection, agreement, and trust. We perturb the content and tone of reasoning chains (Fig.~\ref{fig:perturbation_pipeline}b, c) of the model judgment on multimodal social scenarios (Fig.~\ref{fig:perturbation_pipeline}a) to measure their effects on user responses.

\smallskip \noindent\textbf{Scenario Selection.}
We use the MORALISE benchmark~\citep{lin2025moralisestructuredbenchmarkmoral} to draw multimodal social scenarios, each consisting of an image and a text annotated with a binary label (\textsc{acceptable} vs.\ \textsc{unacceptable}). Unlike domains such as math or logic, MORALISE emphasizes \emph{socially grounded, multimodal situations} where judgments can be ambiguous and explanations can be persuasive even when imperfect. We exclude scenarios containing graphic violence or hate. From the remaining set, we sample 200 \textsc{Acceptable} and 200 \textsc{Unacceptable} scenarios, yielding 400 scenarios. For each scenario, we construct two MORALISE-relative verdict conditions: a \emph{correct} verdict, where the model's final judgment matches the MORALISE label, and an \emph{incorrect} verdict, with respect to the MORALISE label. 
Since our main question concerns how reasoning errors and confidence tone affect user scrutiny, we average over verdict conditions in the main analysis and report verdict-condition-stratified results in \S~\ref{app:judgment-correcntess-results}. Each scenario is assigned one reasoning-error condition at the scenario level. Crossing 400 scenarios with two verdict conditions yields 800 stimuli.

\smallskip \noindent\textbf{Study Setup.} 
We employ a fully between-subjects design. Each scenario is assigned one reasoning-error condition (\textsc{Clean}, \textsc{Omission}, \textsc{Contradiction}, or \textsc{Hallucination}) at the scenario level, and each stimulus is independently assigned one confidence tone (\S\ref{sec:confidence-tones}). Each participant is assigned to a single stimulus---one image--scenario pair, one explanation, one verdict, and one tone---and responds to three measures in a fixed order: reasoning-error detection, verdict agreement, and trust/willingness to rely on the model. This design avoids carry-over effects that could arise from exposing the same participant to multiple error types or tones~\citep{buccinca2021trust}. We recruit $800$ AMT workers, compensated at \$12--15/hour.

\subsection{Correctness of Reasoning Chains}
\label{sec:perturbations}
\smallskip \noindent\textbf{Clean Reasoning Chains.} We first construct \emph{faithful and coherent reasoning chains} that describe the visual and textual context without contradiction or hallucinations and flow logically from observations to a moral conclusion, which serve as a control for studying user trust. 
We use GPT-4V~\citep{openai2024gpt4technicalreport} with a prompt that enforces grounded description and coherent step-by-step reasoning (see \S\ref{app:prompt_example}). This can be applied to reasoning chains produced by any CoT-capable model.  
To confirm that these chains are clean---i.e., free of the error types in our taxonomy---we apply a two-stage verification. First, GPT-5.2~\citep{openai2025gpt52systemcard} screens each chain against our error taxonomy, flagging any omission, contradiction, or hallucination. Then, the chains are verified by two independent annotators over all clean chains for each scenario (Cohen's $\kappa$=0.87). We also validate robustness with Gemini-2.5 Pro~\citep{comanici2025gemini}, obtaining high cross-judge agreement ($91.5\%$). Since perturbations are introduced by design, the judge's role is simply to verify that the intended error is present and no unintended errors were introduced, a far more constrained task than open-ended evaluation. 

\paragraph{Perturbed Reasoning Chains.}
From the clean chains, we induce errors of three types to create flawed reasoning chains. These capture \emph{incompleteness} (\Omit{}), \emph{inconsistency} (\Contr{}), and \emph{ungrounded invention} (\Hall{}), observing taxonomies proposed in faithfulness metrics \citep[e.g., ROSCOE;][]{golovneva2023roscoesuitemetricsscoring}, perturbation-based model reasoning evaluations~\citep{lanham2023measuringfaithfulnesschainofthoughtreasoning}, and analyses of explanation failures in multimodal contexts~\citep{manuvinakurike2025thoughtsthinkingreconsideringexplanatory}.
\begin{itemize}[itemsep=-4pt,topsep=0pt]
    \item \textbf{\Omit{}}: Skipping critical premises or inferential steps, leading to incomplete but seemingly plausible chains~\citep{lanham2023measuringfaithfulnesschainofthoughtreasoning, golovneva2023roscoesuitemetricsscoring}. In multimodal settings, omissions often occur when perceptual details are ignored.  
    \item \textbf{\Contr{}}: Introducing internal inconsistencies or logical conflicts, such as negating an earlier claim or reaching a conclusion unsupported by prior steps~\citep{golovneva2023roscoesuitemetricsscoring, manuvinakurike2025thoughtsthinkingreconsideringexplanatory}.
    \item \textbf{\Hall{}}: Fabricating details that are not grounded in the input (image and text scenario), such as adding nonexistent entities or causal links~\citep{Ji_2023, park2025halloc}.  
\end{itemize}
With the verified \textsc{Clean} chains, we construct localized errors via controlled perturbations, which lets us isolate the causal effect of each error type on trust---model outputs in the wild contain mixed, confounded patterns (\S\ref{sec:5_profilingvlms}). We generate \textsc{Omission}, \textsc{Contradiction}, and \textsc{Hallucination} variants using standardized edit rules (Alg.~\ref{alg:perturb} in \S\ref{app:stimulus-construction-verification}), and re-run the judge to confirm that (i) the target error is present and (ii) no additional error types are introduced (\S\ref{app:val-perturbed-chains}). 
Finally, two annotators confirm the presence of the intended error type in \textbf{95.8\%} of cases.
Unintended secondary errors occur in 3.3\% of chains, most commonly involving borderline omissions following contradiction edits. Overall coherence is rated as preserved in \textbf{97.5\%} of cases. Inter-annotator agreement is high (Cohen’s $\kappa{=}0.84$). Details to the validations are in  \S~\ref{app:perturb_validation}.

\subsection{Confidence Tones}
\label{sec:confidence-tones}
Confidence tone is assigned to each stimulus at random, independently of its error condition, in one of three stylistic variants rendered by corresponding markers: \textbf{hedged} (\emph{low certainty} induced by modals such as \emph{might, may, could}, adverbs such as \emph{possibly, perhaps}, and verbs such as \emph{seems, appears}), \textbf{neutral} (\emph{medium certainty} induced with markers like \emph{likely, suggests}), and \textbf{confident} (high certainty induced by adverbs like \emph{definitely}, adjectives like \emph{clear, obvious}, and phrases like \emph{without doubt, it is evident that}). These are well-established cues for trust calibration, with hedged tones lowering perceived trust and confident tones inflating trust even for wrong answers~\citep{hyland1998hedging, 
pavlick-tetreault-2016-empirical, mielke2022reducing, zhou2024relying, 
sharma2024suggestthathumantrust, Sokol_2024}.
Our lexical substitution approach isolates this single dimension to enable clean causal attribution; we verify that tone edits do not introduce surface-level confounds (e.g., length) via checks in \S~\ref{app:tone_check}.
\subsection{Measures for Trust Calibration}
\label{sec:trust-measures}
Our approach operationalizes trust as a multi-layered construct rather than a single attitude. \citet{Lee2004TrustIA} describe how trust in automation operates as a causal chain from beliefs to attitudes to intentions to behaviors, and these layers can diverge: self-reported trust attitudes do not reliably predict behavioral engagement~\citep{visser2023trustdistrustappropriatereliance, korber2018}, and behavioral measures capture dimensions that attitudinal measures miss~\citep{buccinca2021trust}. \citet{kohn2021} also note that behavioral measures better reflect actual engagement. We therefore triangulate trust across three measures, separating \emph{behavioral} signals (what users do in response to a reasoning chain) from \emph{perceived} trust (what users report feeling), with each measure capturing a different layer of this causal chain (Example stimuli are in \S~\ref{sec:appendix:study_examples}):

\begin{enumerate}[leftmargin=*, itemsep=2pt, topsep=2pt, parsep=0pt]
\item \textbf{Error Detection (binary)} \textit{``Does the reasoning contain a factual or logical error?''}: A behavioral proxy for critical engagement with the reasoning. Behavioral measures of critical engagement diverge meaningfully from self-reported trust~\citep{visser2023trustdistrustappropriatereliance, buccinca2021trust}; by asking participants to explicitly flag errors, we capture \emph{reasoning-critical scrutiny} that purely attitudinal measures can miss.\footnote{This binary probe captures spontaneous, low-friction scrutiny—whether a reader flags a problem under conditions resembling ordinary use, where explanations are often skimmed. It does not establish exhaustive or sentence-level review, and we do not claim it does. We report response-time robustness checks in \S~\ref{appendix:study_details} showing that detection rates are stable when fast responses are excluded.}
\item \textbf{Agreement with Judgment (7-pt Likert)} \textit{``How much do you agree with the model's final judgment?''}: A behavioral proxy at the outcome level, indicating whether participants endorse the model's decision. Agreement is widely used in explainability studies~\citep{Lee2004TrustIA, bansal2021does, vereschak2024trust} as an indicator of how users respond to AI outputs.
\item \textbf{Self-Reported Trust (7-pt Likert)} \textit{``Would you trust the system to make similar judgments?''}: A perceived-trust measure, consistent with HCI and XAI research using subjective ratings of willingness to delegate~\citep{lago2025evaluatingexplainabilityframeworksystematic, visser2023trustdistrustappropriatereliance, Atf_2025}. Single-item trust measures are common in human--AI decision-making studies~\citep{lai2023towards}.
\end{enumerate}
Together, these measures triangulate trust across complementary layers: error detection captures \emph{behavioral scrutiny} of the reasoning, agreement captures \emph{behavioral endorsement} of the verdict, and self-reported trust captures \emph{perceived trustworthiness} of the system.
We treat a flawed reasoning chain as something users should detect and discount. 
Explanations not only justify correct outputs but also surface reasons to distrust flawed ones~\citep{ribeiro-etal-2016-trust, jacovi2021formalizingtrustartificialintelligence}, and appropriate reliance is the ability to tell sound output from flawed ones and act on that difference~\citep{Schemmer_2023, Lee2004TrustIA}. Miscalibration is the failure of this coupling: agreeing with and trusting a chain because its error went unnoticed. This is a standard we evaluate against, not a quantity we compute: each measure is analyzed separately, and miscalibration is measured as the co-occurrence of a missed error with high reliance (\S\ref{app:miscalibration_rates}). 

\section{Detecting Errors in Controlled Reasoning Chains}
Our analyses address three guiding questions:
(1) \emph{How do flawed reasoning chains impact user trust in model judgments?}
(2) \emph{How does certainty in tone affect sensitivity to reasoning correctness?} 
(3) \emph{How well do models detect the same errors in reasoning chains?}

\subsection*{Q1. How do flawed reasoning chains impact user trust in model judgments?}
\label{sec:results_q1}
\begin{table*}[t]
  \centering
  \small
  \setlength{\tabcolsep}{4pt}
  \begin{tabular}{lccc}
    \toprule
    \textbf{Perturbation Type} & \textbf{Detection (\%)} & \textbf{Agreement (1--7)} & \textbf{Trust (1--7)} \\
    \midrule
    \Clean{} & \cellcolor{green!10} 15.5 \scriptsize{[9.6, 19.8]} & \cellcolor{cyan!40} 4.21 \scriptsize{[3.93, 4.49]} & \cellcolor{red!40} 3.23 \scriptsize{[2.99, 3.47]} \\
    \Omit{}  & \cellcolor{green!30} 39.4 \scriptsize{[32.6, 46.6]} & \cellcolor{cyan!30} 4.00 \scriptsize{[3.77, 4.23]} & \cellcolor{red!30} 2.86 \scriptsize{[2.67, 3.06]} \\
    \Contr{} & \cellcolor{green!60} 69.0 \scriptsize{[62.6, 75.4]} & \cellcolor{cyan!20} 3.52 \scriptsize{[3.30, 3.73]} & \cellcolor{red!20} 2.33 \scriptsize{[2.15, 2.51]} \\
    \Hall{}  & \cellcolor{green!55} 66.3 \scriptsize{[59.5, 72.6]} & \cellcolor{cyan!20} 3.38 \scriptsize{[3.14, 3.62]} & \cellcolor{red!15} 2.22 \scriptsize{[2.03, 2.41]} \\
    \bottomrule
  \end{tabular}
  \caption{\textbf{Main outcomes across conditions.} Strong agreement--trust coupling and clear error-type differences emerge, with \emph{omissions} the hardest to detect. Color intensity encodes the relative magnitude of each metric: darker green = $\uparrow$ detection accuracy, darker blue = $\uparrow$ agreement, darker red = $\uparrow$ perceived trust.}
  \label{tab:by-error-main}
\end{table*}

\begin{figure}[t]
  \centering
  \includegraphics[width=0.95\linewidth]{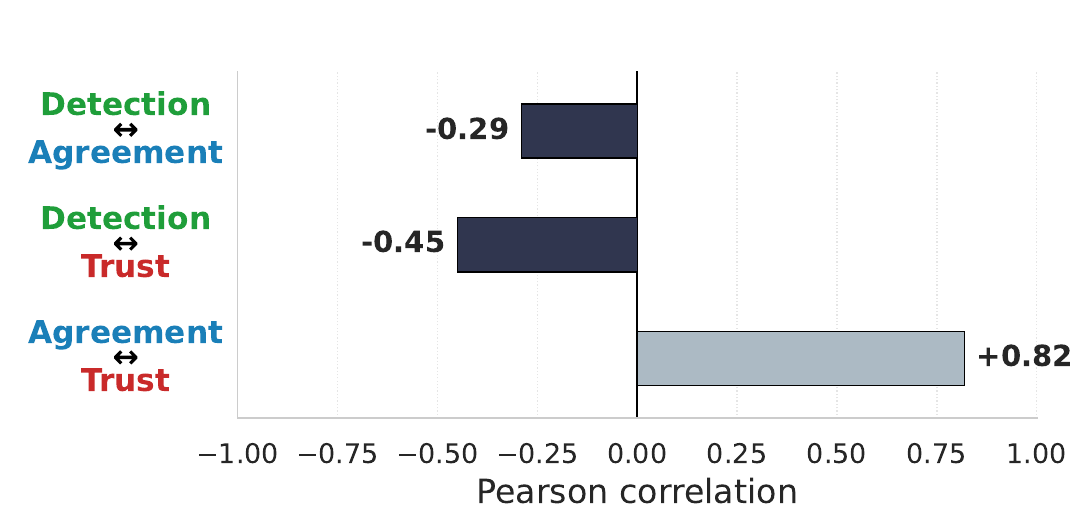}
  \caption{\textbf{Overall cross-measure correlations.} Negative correlations indicate that higher agreement and trust coincide with reduced error detection.}
  \label{fig:cross-overall-main}
\end{figure}

Table~\ref{tab:by-error-main} summarizes participants' responses across error types, showing clear contrasts in detection, agreement, and trust.
We find that \textbf{users' agreement with the model's output is the most dominant factor influencing trust, even when the reasoning is flawed}.
Fig.~\ref{fig:cross-overall-main} quantifies these patterns: agreement and trust are strongly correlated ($r{=}{+}0.82$), and both are inversely related to detection (agreement: $r{=}{-}0.29$; trust: $r{=}{-}0.45$). These negative associations indicate that \textit{higher agreement and trust coincide with lower error detection}, suggesting users' reported trust primarily reflects alignment with the model's verdict rather than the fidelity of its reasoning. 

Error types differ in how easily they are detected. The mixed-effects model in Table~\ref{tab:stats} shows that \Contr{} ($\beta{=}{+}2.59$) and \Hall{} ($\beta{=}{+}2.47$) nearly double the detection gain of \Omit{} ($\beta{=}{+}1.33$), and significantly reduce both agreement and trust. \Omit{}, by contrast, leaves agreement and trust largely unchanged; despite being logically incomplete, omissions go unnoticed in nearly 60\% of cases and still elicit high agreement (4.00) and trust (2.86), only marginally below \textsc{Clean} (4.21 / 3.23). \looseness=-1

\begin{table}[h]
  \centering
  \small
  \resizebox{0.95\linewidth}{!}{
  \begin{tabular}{lccc}
  \toprule
   & \textbf{Detection} & \textbf{Agreement} & \textbf{Trust} \\
  \midrule
  \multicolumn{4}{l}{\textit{Coefficients ($\beta$, vs.\ \Clean{} / Neutral)}} \\
  \addlinespace[2pt]
  \Omit{}        & $+1.33^{\dagger}$ & $-0.22^{\diamond}$           & $-0.37^{\dagger}$           \\
  \Contr{}       & $+2.59^{\dagger}$ & $-0.70^{\dagger}$ & $-0.90^{\dagger}$ \\
  \Hall{}        & $+2.47^{\dagger}$ & $-0.83^{\dagger}$ & $-1.01^{\dagger}$ \\
  Confident Tone & $-0.49^{*}$           & $-0.08$           & $-0.00$           \\
  \midrule
  \multicolumn{4}{l}{\textit{ANOVA test statistics}} \\
  \addlinespace[2pt]
  Error type main      & $159.64^{\dagger}$ & $10.09^{\dagger}$ & $20.94^{\dagger}$ \\
  Tone main            & $7.56^{*}$         & $0.44$            & $0.65$            \\
  \bottomrule
  \end{tabular}}
  \caption{\textbf{Mixed-effects model coefficients and ANOVA results.}
  Detection fit by logistic regression. Agreement and Trust by OLS on 1--7 ratings.
  $^{*}p{<}.05$, $^{\diamond}p{<}.01$, $^{\dagger}p{<}.001$ (Holm-corrected).}
  \label{tab:stats}
\end{table}

\smallskip \noindent\textbf{Discussion.}
Our findings align with prior work showing that users often evaluate explanations based on outcome agreement or fluency rather than critical scrutiny of validity ~\citep{jacovi2021formalizingtrustartificialintelligence, lanham2023measuringfaithfulnesschainofthoughtreasoning}.
We extend this literature by confirming a strong association between agreement and reduced scrutiny; participants who align with the model's judgment are less likely to examine the reasoning that led there. This effect holds for both clean and perturbed chains. The low omission-type error detection rate may partly reflect an \emph{enthymematic} effect~\citep{macagno2013dialogical}: if users perceive omitted steps as too obvious to mention, they accept incomplete chains without scrutinizing whether the unstated premise is warranted, a risk that holds whether the omission is a genuine reasoning failure or efficient compression. This may be amplified in multimodal settings, where visual context further enhances plausibility~\citep{delaunay2025impact}.
Thus, we report a failure of \emph{error-sensitive trust}: users continue to report trust on a model even when they fail to notice flaws in its reasoning; such misplaced trust manifests as cases where users fail to detect reasoning error but nevertheless show high self-reported trust.
\begin{table}[t]
\centering
\small
\begin{tabular}{lcc}
\toprule
Flawed trials & Agreement (1--7) & Trust (1--7) \\
\midrule
Error detected & 3.30 & 1.99 \\
Error missed   & 4.08 & 3.12 \\
\addlinespace[2pt]
\textit{\textsc{Clean} reference} & \textit{4.21} & \textit{3.23} \\
\midrule
\multicolumn{3}{l}{\textit{Missed error \emph{and} high reliance (\%)}} \\
\quad \textsc{Omission}      & \multicolumn{2}{c}{27.5} \\
\quad \textsc{Contradiction} & \multicolumn{2}{c}{14.5} \\
\quad \textsc{Hallucination} & \multicolumn{2}{c}{13.0} \\
\bottomrule
\end{tabular}
\caption{\textbf{Missed errors sustain reliance.} Top: on flawed trials, participants who flagged the error reported markedly lower agreement and trust than those who did not; the latter remain near the \textsc{Clean} baseline. Bottom: proportion of trials combining a missed error with high reliance (Agreement $\geq5$ or Trust $\geq5$), by error type. The detection split is measured, not assigned.}
\label{tab:detected-missed}
\end{table}

\smallskip \noindent\textbf{Missed errors sustain reliance.} The preceding analyses aggregate over participants. To test whether reliance is coupled to scrutiny at the participant level, we split flawed trials by whether the participant flagged an error (Table~\ref{tab:detected-missed}). Participants who detected the error reported substantially lower agreement (3.30) and trust (1.99) than those who missed it (4.08 and 3.12) and those who missed it remained statistically indistinguishable from the CLEAN baseline (4.21 / 3.23). The asymmetry is largest for omissions: 27.5\% of omission trials combine a missed error with high reliance, versus 14.5\% for contradictions and 13.0\% for hallucinations. Our findings therefore do not show that agreement causes less attention, but show that when an error goes unnoticed, reliance proceeds as though the chain was free of errors.

\subsection*{Q2. How does certainty in tone affect sensitivity to reasoning correctness?} 

\begin{table}[t]
  \centering
  \resizebox{\linewidth}{!}{%
  \begin{tabular}{llcc}
  \toprule
  \textbf{Tones} & \textbf{Detection (\%)} & \textbf{Agreement (1--7)} & \textbf{Trust (1--7)} \\
  \midrule
  Hedged    & \cellcolor{green!57} 50.4 \scriptsize{[44.0, 56.8]} & \cellcolor{cyan!10} 3.70 \scriptsize{[3.48, 3.92]} & \cellcolor{red!10} 2.56 \scriptsize{[2.38, 2.75]} \\
  \midrule
  Neutral   & \cellcolor{green!60} 47.0 \scriptsize{[40.9, 53.2]} & \cellcolor{cyan!60} 3.85 \scriptsize{[3.62, 4.06]} & \cellcolor{red!50} 2.70 \scriptsize{[2.52, 2.88]} \\
  \midrule
  Confident & \cellcolor{green!10} \textbf{40.5 \scriptsize{[34.6, 46.7]}} & \cellcolor{cyan!34} 3.78 \scriptsize{[3.58, 3.98]} & \cellcolor{red!50} 2.70 \scriptsize{[2.53, 2.89]} \\
  \bottomrule
  \end{tabular}}
  \caption{\textbf{Confidence tone effects.} Numbers show mean with 95\% bootstrap CIs. \textit{Neutral} denotes the baseline wording of each chain \emph{before} tone edits.}
  \label{tab:tone_effects}
  \end{table}

Table~\ref{tab:tone_effects} reports detection, agreement, and trust across tone conditions, averaged across the three error types as reasoning chains---both clean and perturbed---are tone-modulated in identical ways. Across all conditions, \textbf{confident tones consistently suppress error detection; participants are less likely to notice reasoning flaws}, while leaving agreement and self-reported trust largely unchanged.

\begin{figure}[t]
    \centering
    \includegraphics[width=\linewidth]{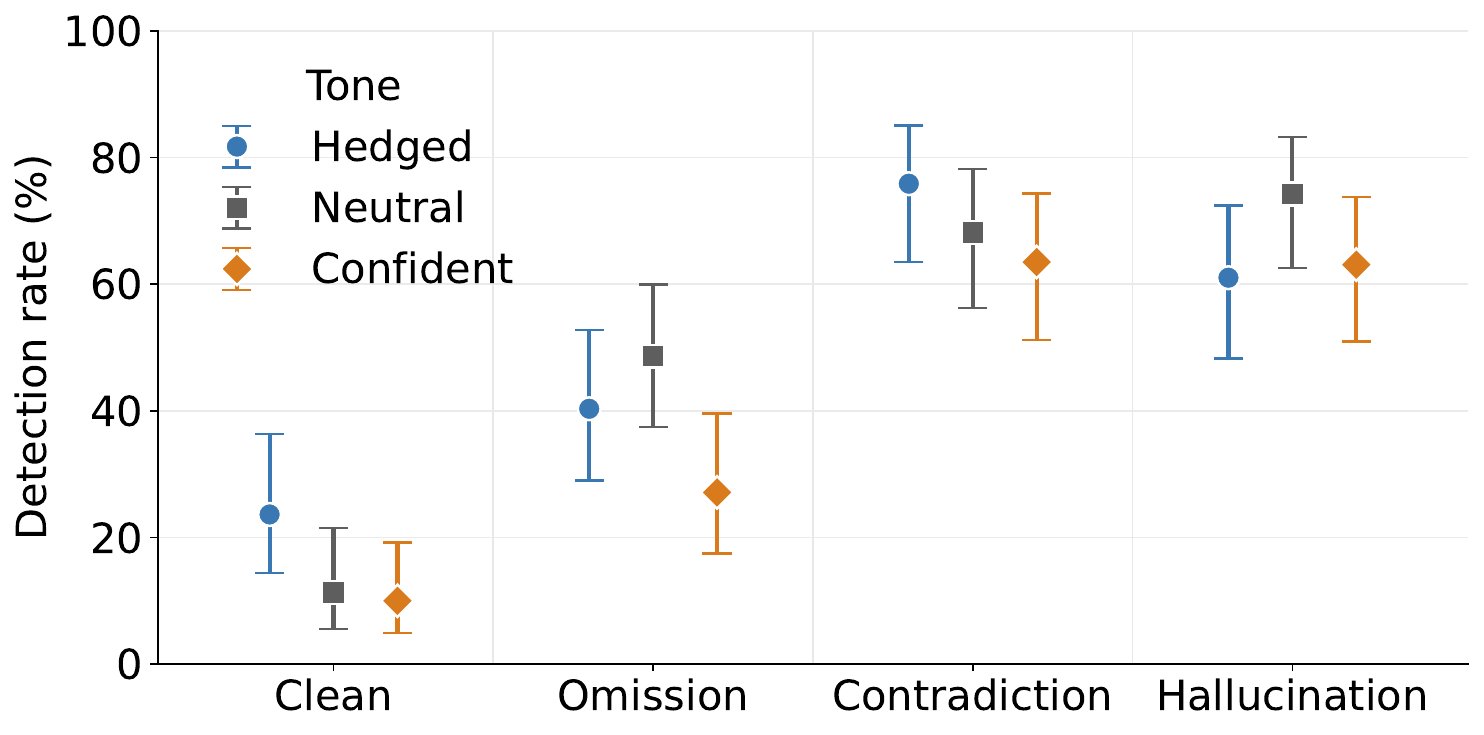}
    \caption{\textbf{Error detection rates by tones}. Points show observed error detections, and vertical error bars indicate 95\% confidence intervals computed separately for each tone $\times$ perturbation-type.}
    \label{fig:tone}
\end{figure}

To confirm this asymmetry, we fit separate mixed-effects models per outcome. Error detection (binary) is fit with logistic mixed-effects regression; agreement and trust (1–7) with linear mixed-effects regression. Fixed effects are reasoning-error condition, expressed-confidence tone, benchmark-relative verdict correctness, and their interactions; reference categories are CLEAN and NEUTRAL. All models include scenario-level random intercepts to absorb item-specific difficulty, so coefficients reflect within-item contrasts rather than differences in scenario ambiguity. Post-hoc contrasts are Holm-corrected.
The results in Table~\ref{tab:stats} show that the tone main effect on detection is significant, with confident framing reducing detection log-odds ($\beta{=}{-}0.49$), whereas tone effects on agreement and trust are null (both $F{<}1$, $p{>}.5$).  Fig.~\ref{fig:tone} nonetheless shows that the effect is most pronounced for \Omit{} errors, and that confident explanations sustain perceived trust---suggesting tone shapes how users evaluate the \emph{process} of reasoning, not just whether they agree with its \emph{outcome}. \looseness=-1

\smallskip \noindent\textbf{Discussion.}
Our results resonate with prior works on linguistic framing showing that confident delivery inflates perceived trustworthiness even when the content is incorrect~\citep{zhou2024relying, zhou2024rel, sharma2024suggestthathumantrust}. 
We extend this line of work by showing that tone effects persist when confidence is expressed within \emph{reasoning} grounded in multimodal evidence, and by separating its effects across our three measures: confident tone suppresses detection while leaving agreement and reported trust unchanged. Users therefore keep relying on outputs whose errors they have not noticed---a dissociation between scrutiny and endorsement that points to a need for explanations balancing informativeness with epistemic humility.

\subsection*{Q3. How well do models detect errors in flawed reasoning chains?}
\label{sec:results_q3}
\begin{figure}[t]
  \centering
  \includegraphics[width=\linewidth]{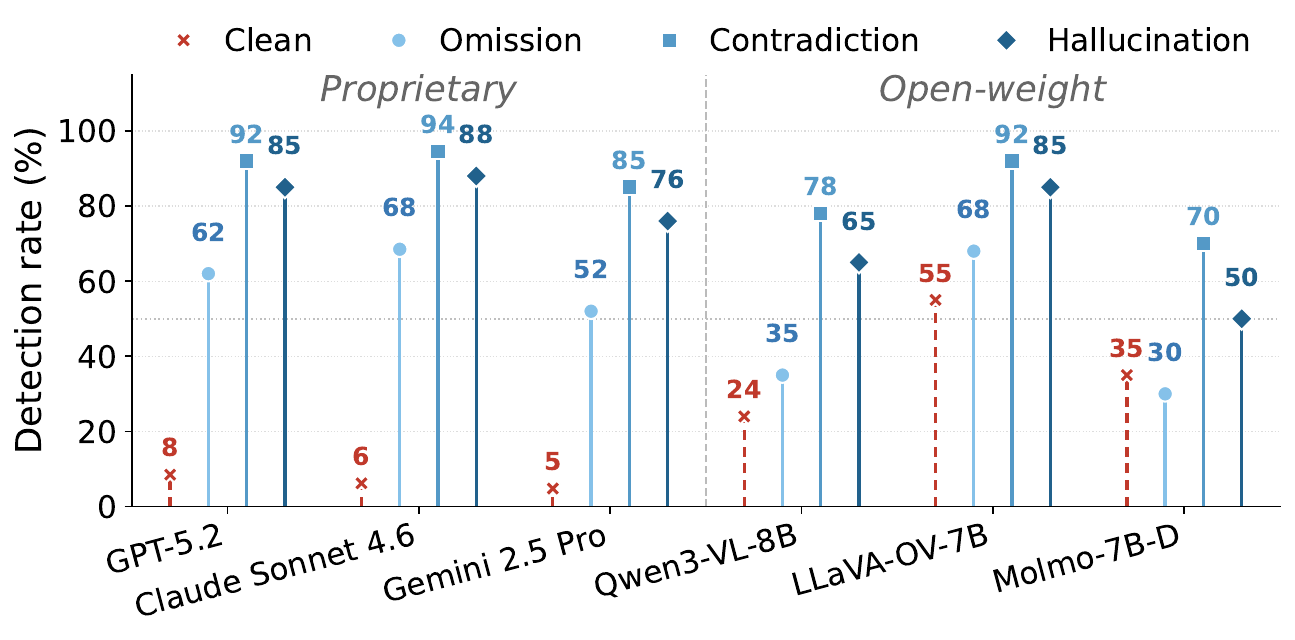}
  \caption{\textbf{Model-side error detection across perturbation types.} \Clean{} bars show false-positive rate ($\downarrow$); the other three show true-positive rate ($\uparrow$) per error type.}
  \label{fig:llm-judge-detection}
\end{figure}

Beyond the human side, we ask whether models themselves can detect the reasoning flaws shown to participants. This question is motivated by the generate-then-verify paradigm, in which LLMs are increasingly deployed as judges of their own or other model's reasoning chains~\citep{weng-etal-2023-large, zheng2023judging}---an approach validated by whether verifier models can flag the same errors as human reviewers. 
We also prompt six VLMs---spanning three closed-source models (\texttt{GPT-5.2}~\citep{openai2025gpt52systemcard}, \texttt{Claude Sonnet 4.6}~\citep{anthropic2026sonnet46}, and \texttt{Gemini 2.5 Pro}~\citep{comanici2025gemini}) and three open-weight models (\texttt{LLaVA-OneVision-7B}~\citep{li2024llava}, \texttt{Qwen3-VL-8B}~\citep{bai2025qwen3}, and \texttt{Molmo-7B-D}~\citep{deitke2025molmo})---with the image, the perturbed reasoning chain, and the final judgment, and ask them to identify whether an error is present and if so, which type. Fig.~\ref{fig:llm-judge-detection} reports false-positive rates on \Clean{} chains and true-positive rates on \Omit{}, \Contr{}, and \Hall{} chains. Mirroring the human-side results, \textbf{\Omit{} remains the hardest error type to detect across all models} (30--68\%), substantially below \Contr{} (70--94\%) and \Hall{} (50--88\%).\looseness=-1

\smallskip \noindent\textbf{Stability of model-side detection.} Single-call detection may understate variance, so we re-run the verifier task with five independent calls per item on a fixed subset (50 chains per condition). Contradictions are correctly identified in 100\% of calls and hallucinations in 95.6\%, while omissions reach only 28.4\%. Item-level stability shows that contradictions are detected in all five calls for every item, whereas only 42.0\% of omissions are detected even once. Omission detection is thus not merely lower on average but unstable across samples.

\smallskip \noindent\textbf{Discussion.}
Omissions being the hardest error for both humans and models point to a shared mechanism: detecting an omission means noticing what is absent, whereas contradictions and hallucinations present a concrete, locatable cue. Spotting missing premises is therefore harder than flagging an inconsistency or a fabrication, for both human readers and model verifiers. This has a direct implication for the generate-then-verify paradigm: model-based verification is useful, but it should not be treated as sufficient across all error types. 

\smallskip \noindent\textbf{Verifier detection depends on how the verifier is asked.} Posing the same question to models and humans keeps the tasks comparable but conflates verifier capability with prompt framing. Across three judge prompts (\S~\ref{app:prompt-sensitivity}), we find that contradiction and hallucination detection is largely prompt-insensitive, while omission detection ranges from 31\% under a generic prompt to 79\% under a taxonomy-based prompt (used for our analysis) at the same 20\% false-positive rate on CLEAN chains, and to 90\% under an omission-targeted prompt at a 50\% false-positive rate. Closing the omission gap through targeted prompting therefore trades sensitivity against specificity.

\section{Error Profiling of VLM Reasoning}
\label{sec:5_profilingvlms}
While our controlled user study isolates effects of specific reasoning flaws, we also ask whether such patterns naturally arise in real-world outputs and reasoning chains, by profiling prevalence of the error and epistemic markers in the same VLMs evaluated in \S~\ref{sec:results_q3}. 

\begin{figure}[t]
    \centering
    \includegraphics[width=\linewidth]{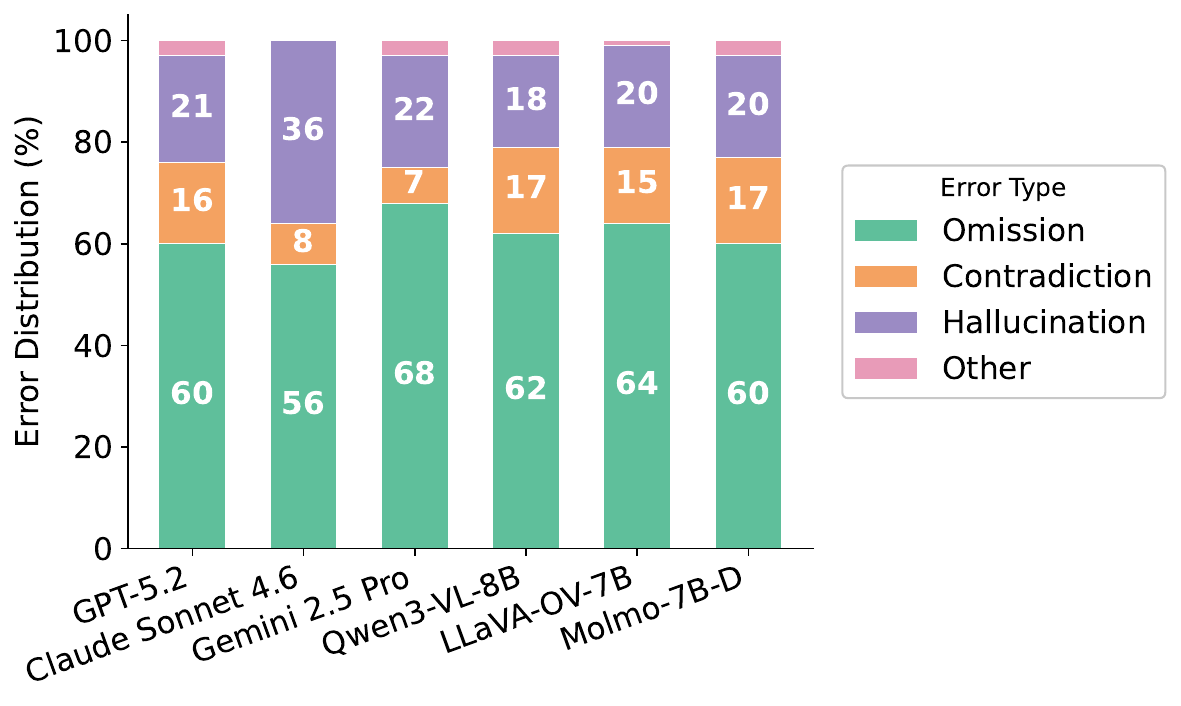}
    \caption{\textbf{In-the-wild error distribution by models.} 
Stacked bar plots of reasoning chain perturbations across six VLMs.}
    \label{fig:model_error_dist}
\end{figure}

\paragraph{Reasoning Error Prevalence Profiling}
For 200 sampled multimodal scenarios, we collect explanations from each model. Each chain is evaluated using GPT-5.2, following the same error taxonomy as in our user study (\S\ref{sec:perturbations}). To validate the reliability of these automatic judgments, two annotators independently review a stratified subset of chains across models and templates, showing 92\% agreement (Cohen’s $\kappa{=}0.86$) for error profiling and 94\% agreement (Cohen’s $\kappa{=}0.89$) for tone profiling. As Fig.~\ref{fig:model_error_dist} shows, omissions are the dominant error type across models, accounting for roughly 56--68\% of profiled chains, while contradictions and hallucinations occur less frequently. 

To further test whether these rates are artifacts of our elicitation prompt, we re-ran the profiling with paraphrased and strictly grounded prompts, shifting GPT-5.2's judged error rate from 23.0\% to 26.0\% and 19.0\% (\S~\ref{app:prompt-sensitivity}): errors persist under all three, including a prompt that explicitly instructs the model to stay grounded and hedge under uncertainty. This suggests that these distributions are a characterization of reasoning under CoT prompting practice rather than prompt-invariant intrinsic error rates. Linking these distributions to human sensitivity (\S\ref{sec:results_q1}), we observe a clear prevalence--detectability gap: \Omit{} errors are the least detectable by humans yet dominate in practice.

\begin{figure}[t]
    \centering
    \includegraphics[width=\linewidth]{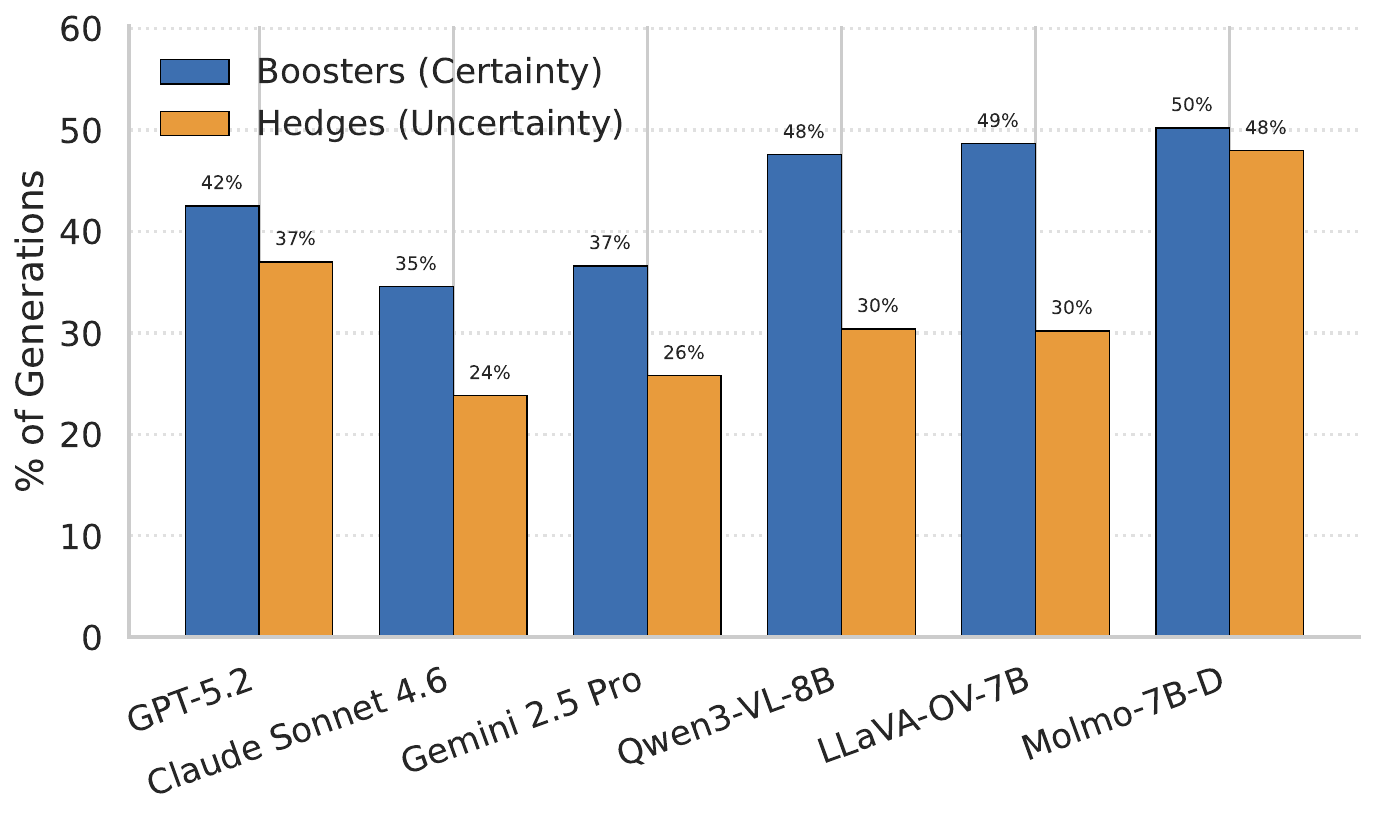}
    \caption{\textbf{Epistemic marker usage across models.} Strengtheners (boosters) are far more common than weakeners (hedges), suggesting a systematic bias toward confident reasoning styles.}
    \label{fig:epistemic_markers}
\end{figure}

\paragraph{Eliciting Epistemic Markers.}
To probe stylistic confidence, we test how often models produce epistemic markers of certainty or uncertainty in reasoning chains. Following \citet{zhou2024relying}, we use three prompting templates: (1) CoT-only (``Explain your reasoning step by step''), (2) Certainty-only (``Provide your judgment with an explicit certainty level''), and (3) CoT+Certainty (``Explain step by step, explicitly marking certainty or uncertainty''). From 50 samples per model per template, we evaluate outputs using an LLM-as-a-Judge (GPT-5.2) to identify \emph{boosters} (certainty markers, e.g., \textit{definitely}) and \emph{hedges} (uncertainty markers, e.g., \textit{might}). 
Across all models (Fig.~\ref{fig:epistemic_markers}), boosters appear in roughly 35--50\% of generations, while hedges appear in 24--48\%. Boosters are more frequent than hedges, though the gap varies across models.

\paragraph{Discussion}
These in-the-wild analyses reveal two compounding risks. First, omissions of steps are both frequent (especially in closed-source models) and under-detected by humans, making them a hidden driver of overtrust. Second, models overuse certainty markers relative to hedges, biasing their explanations toward confident reasoning styles. This combination, frequent omissions paired with confident delivery, suggests that reasoning chains can be prone to fostering blind trust and amplifying miscalibration in model explanations.

\section{Conclusion}
We introduce a framework for controlled experiments to examine how flawed reasoning chains and confidence tones shape user trust in moral judgments and explanations by VLMs. By systematically perturbing reasoning correctness (\Omit{}, \Contr{}, \Hall{}) and manipulating tones, we triangulate miscalibrated trust through error detection, agreement, and self-reported trust. We show that users often overlook reasoning flaws when they agree with the final outcome, with \Omit{}s emerging as both the most frequent error in real model outputs and the hardest to detect. 

We also find that confident tones suppress scrutiny while leaving self-reported trust intact, so reliance is sustained even when the underlying reasoning is flawed.
Together, these findings highlight a prevalence–detectability gap in real-world models: explanation styles, both in structure and in tone, can actively shape overtrust rather than passively convey reasoning. 
Most explainability research has focused on the model side, improving the faithfulness, calibration, and interpretability of generated explanations~\citep{lyu-etal-2023-faithful, Chen2025ReasoningMD, turpin2023language}. Our findings suggest the user side deserves comparable attention: even well-formed reasoning chains fail to support calibrated trust when users do not critically scrutinize them. More broadly, our results have implications for monitorability research~\citep{Korbak2025ChainOT, Meek2025MeasuringCM}, which has largely framed oversight as a question of whether internal reasoning faithfully reflects model behavior. Our findings suggest a complementary front: external scrutiny of reasoning is driven less by chain quality than by surface signals like outcome agreement and tone, so even faithful reasoning chains may go unaudited when the verdict feels right. Outcome-level signals such as user–model disagreement could therefore be useful for closer reasoning audits. We manipulate error type and expressed confidence; whether epistemic humility or disagreement-triggered audits improve detection is testable within it---the same stimuli with an added audit-prompt condition---but untested here. We therefore offer these as design implications rather than validated interventions.

\section{Limitations} 
\label{sec:limitations}
While our study provides new insights into how reasoning chains and confidence tone shape user trust, several limitations remain.
First, our design relies on systematically constructed perturbations rather than reasoning chains generated in-the-wild. Although this lets us isolate error types and control for correctness, model outputs often contain mixed or subtler error patterns that our taxonomy may not fully capture. In particular, some omission errors may have been perceived by participants as "too obvious to mention" rather than as genuine reasoning failures, potentially leading to under-detection. Future work could examine whether different classes of omissions (missing inferential steps versus omitted but socially implicit information) have qualitatively distinct effects on user trust.

Second, our study focuses on a specific domain of model usage, multimodal moral judgment. This setting is well-suited for probing trust because it involves intuitive, socially grounded reasoning to which most participants can relate, yet it may limit generalizability. Extending the design to text-only domains, where the perceptual layer is absent, would clarify whether confidence-induced overtrust is driven by the structure and tone of the reasoning chain itself or whether multimodal perception introduces an additional source of uncertainty.

Third, our manipulation of confidence tone uses controlled lexical substitutions (hedges, boosters, neutrals). This ensures internal validity but may not capture the full pragmatic richness of tone in natural dialogue, such as prosody, discourse context, or stylistic variation across languages; studying multimodal delivery (e.g., voice, gesture, or interface framing) remains an open challenge.

Fourth, our model-side detection study (\S\ref{sec:results_q3} Q3) prompts VLM verifiers with the same generic error-detection question given to human participants. While this keeps the human and model tasks comparable, it likely understates model capability: a verifier explicitly asked whether a necessary premise is missing may detect omissions far more reliably than what we observe under generic prompting. Our model-side rates should therefore be read as a floor rather than a ceiling, and pairing model-based checks with targeted prompts or human validation is a concrete direction for future work.

Finally, our measures of trust---detection, agreement, and self-report---capture complementary aspects of user engagement but are not exhaustive. We do not measure downstream reliance in consequential tasks (e.g., decision-making or delegation), which would provide stronger evidence of real-world impact.

Overall, these limitations reflect the tradeoff between experimental control and ecological validity. We see our work as a step toward a more comprehensive understanding of trust calibration, and encourage future work to test our framework across broader domains, user populations, and multimodal delivery settings.

\section{Ethical Considerations}

\paragraph{Participant Safety and Consent.}
All user studies were conducted under IRB approval from our institution. Participants were recruited via the Amazon Mechanical Turk platform and provided informed consent prior to participation. The moral scenarios used in this study were drawn from publicly available datasets (MORALISE~\citep{lin2025moralisestructuredbenchmarkmoral}) and were screened to exclude depictions of graphic violence, hate speech, or personally identifiable information. Participants were also informed that they could withdraw at any time without penalty.
All collected responses were anonymized and stored securely. No personally identifiable information was collected beyond basic demographics (e.g., age group, country). Data were used exclusively for research purposes and will be released only in an aggregate form to ensure participant privacy.  

\paragraph{Model Outputs and Potential Harms.}
The study involves evaluating and manipulating outputs from VLMs. While reasoning chains were filtered for appropriateness, some text might contain implicit biases or morally ambiguous content. 

\paragraph{Broader Impacts and Responsible Use.}
Our findings highlight that explanations, especially well-structured reasoning chains, can inadvertently foster blind trust. We caution against deploying unfaithful or overconfident reasoning outputs in high-stakes contexts such as healthcare, education, or legal decision support. Developers and practitioners should treat explanatory reasoning as a communication feature that requires careful calibration and user testing, not as a guarantee of model reliability. 

\section{Acknowledgments}

We thank Mingqian Zheng for her valuable feedback. This work was supported by the SNU-Global Excellence Research Center establishment project. This research was supported by Basic Science Research Program through the National Research Foundation of Korea (NRF) funded by the Ministry of Education (RS-2023-00274280). This work was supported by the IITP(Institute of Information \& Communications Technology Planning \& Evaluation)-ITRC (Information Technology Research Center) grant funded by the Korea government (Ministry of Science and ICT) (IITP-2025-RS-2024-00437633). This work was supported by the Institute of Information \& Communications Technology Planning \& Evaluation(IITP) grant funded by the Korea government (MSIT) (RS-2025-25442338, AI star Fellowship Support Program (Seoul National Univ.)). This work was supported by Institute of Information \& communications Technology Planning \& Evaluation (IITP) grant funded by the Korea government (MSIT) (No.~RS-2021-II211343, Artificial Intelligence Graduate School Program (Seoul National University)).

\bibliography{latex/publications_cleaned_2026-08-27}

\begin{thebibliography}{58}
\providecommand{\natexlab}[1]{#1}

\bibitem[{Agarwal et~al.(2024)Agarwal, Tanneru, and Lakkaraju}]{agarwal2024faithfulness}
Chirag Agarwal, Sree~Harsha Tanneru, and Himabindu Lakkaraju. 2024.
\newblock \href {https://arxiv.org/abs/2402.04614} {Faithfulness vs. plausibility: On the (un)reliability of explanations from large language models}.
\newblock \emph{Preprint}, arXiv:2402.04614.

\bibitem[{{Anthropic}(2026)}]{anthropic2026sonnet46}
{Anthropic}. 2026.
\newblock Introducing claude sonnet 4.6.
\newblock \url{https://www.anthropic.com/news/claude-sonnet-4-6}.
\newblock Accessed: 2026-05-11.

\bibitem[{Atf and Lewis(2025)}]{Atf_2025}
Zahra Atf and Peter~R. Lewis. 2025.
\newblock \href {https://doi.org/10.1109/tts.2025.3558448} {Is trust correlated with explainability in ai? a meta-analysis}.
\newblock \emph{IEEE Transactions on Technology and Society}.

\bibitem[{Bai et~al.(2025)Bai, Cai, Chen, Chen, Chen, Cheng, Deng, Ding, Gao, Ge, Ge, Guo, Huang, Huang, Huang, Hui, Jiang, Li, Li, Li, Li, Lin, Lin, Liu, Liu, Liu, Liu, Liu, Liu, Lu, Luo, Lv, Men, Meng, Ren, Ren, Song, Sun, Tang, Tu, Wan, Wang, Wang, Wang, Wang, Xie, Xu, Xu, Xu, Yang, Yang, Yang, Yang, Yu, Zhang, Zhang, Zhang, Zheng, Zhong, Zhou, Zhou, Zhou, Zhu, and Zhu}]{bai2025qwen3}
Shuai Bai, Yuxuan Cai, Ruizhe Chen, Keqin Chen, Xionghui Chen, Zesen Cheng, Lianghao Deng, Wei Ding, Chang Gao, Chunjiang Ge, Wenbin Ge, Zhifang Guo, Qidong Huang, Jie Huang, Fei Huang, Binyuan Hui, Shutong Jiang, Zhaohai Li, Mingsheng Li, and 45 others. 2025.
\newblock \href {https://arxiv.org/abs/2511.21631} {Qwen3-vl technical report}.
\newblock \emph{Preprint}, arXiv:2511.21631.

\bibitem[{Bansal et~al.(2021)Bansal, Wu, Zhou, Fok, Nushi, Kamar, Ribeiro, and Weld}]{bansal2021does}
Gagan Bansal, Tongshuang Wu, Joyce Zhou, Raymond Fok, Besmira Nushi, Ece Kamar, Marco~Tulio Ribeiro, and Daniel Weld. 2021.
\newblock \href {https://api.semanticscholar.org/CorpusID:220128138} {Does the whole exceed its parts? the effect of ai explanations on complementary team performance}.
\newblock In \emph{Proceedings of the 2021 CHI conference on human factors in computing systems}, pages 1--16.

\bibitem[{Basoah et~al.(2025)Basoah, Chechelnitsky, Long, Reinecke, Zerva, Zhou, Díaz, and Sap}]{Basoah_2025}
Jeffrey Basoah, Daniel Chechelnitsky, Tao Long, Katharina Reinecke, Chrysoula Zerva, Kaitlyn Zhou, Mark Díaz, and Maarten Sap. 2025.
\newblock \href {https://doi.org/10.1145/3715275.3732045} {Not like us, hunty: Measuring perceptions and behavioral effects of minoritized anthropomorphic cues in llms}.
\newblock In \emph{Proceedings of the 2025 ACM Conference on Fairness, Accountability, and Transparency}, FAccT ’25, page 710–745. ACM.

\bibitem[{Beck et~al.(2025)Beck, Eckman, Kern, and Kreuter}]{beck2025biasloophumansevaluate}
Jacob Beck, Stephanie Eckman, Christoph Kern, and Frauke Kreuter. 2025.
\newblock \href {https://arxiv.org/abs/2509.08514} {Bias in the loop: How humans evaluate ai-generated suggestions}.
\newblock \emph{Preprint}, arXiv:2509.08514.

\bibitem[{Bu{\c c}inca et~al.(2021)Bu{\c c}inca, Malaya, and Gajos}]{buccinca2021trust}
Zana Bu{\c c}inca, Maja~Barbara Malaya, and Krzysztof~Z. Gajos. 2021.
\newblock \href {https://doi.org/10.1145/3449287} {To trust or to think: Cognitive forcing functions can reduce overreliance on ai in ai-assisted decision-making}.
\newblock \emph{Proc. ACM Hum.-Comput. Interact.}, 5(CSCW1).

\bibitem[{Chen et~al.(2025)Chen, Benton, Radhakrishnan, Uesato, Denison, Schulman, Somani, Hase, Wagner, Roger, Mikulik, Bowman, Leike, Kaplan, and Perez}]{Chen2025ReasoningMD}
Yanda Chen, Joe Benton, Ansh Radhakrishnan, Jonathan Uesato, Carson~E. Denison, John Schulman, Arushi Somani, Peter Hase, Misha Wagner, Fabien Roger, Vladimir Mikulik, Sam Bowman, Jan Leike, Jared Kaplan, and Ethan Perez. 2025.
\newblock \href {https://arxiv.org/abs/2505.05410} {Reasoning models don't always say what they think}.
\newblock \emph{ArXiv}, abs/2505.05410.

\bibitem[{Comanici et~al.(2025)Comanici, Bieber, Schaekermann, Pasupat, Sachdeva, Dhillon, Blistein, Ram, Zhang, Rosen et~al.}]{comanici2025gemini}
Gheorghe Comanici, Eric Bieber, Mike Schaekermann, Ice Pasupat, Noveen Sachdeva, Inderjit Dhillon, Marcel Blistein, Ori Ram, Dan Zhang, Evan Rosen, and 1 others. 2025.
\newblock \href {https://arxiv.org/abs/2507.06261} {Gemini 2.5: Pushing the frontier with advanced reasoning, multimodality, long context, and next generation agentic capabilities}.
\newblock \emph{arXiv preprint arXiv:2507.06261}.

\bibitem[{DeepSeek-AI et~al.(2025)DeepSeek-AI, Guo, Yang, Zhang, Song, Zhang, Xu, Zhu, Ma, Wang, Bi, Zhang, Yu, Wu, Wu, Gou, Shao, Li, Gao, Liu, Xue, Wang, Wu, Feng, Lu, Zhao, Deng, Zhang, Ruan, Dai, Chen, Ji, Li, Lin, Dai, Luo, Hao, Chen, Li, Zhang, Bao, Xu, Wang, Ding, Xin, Gao, Qu, Li, Guo, Li, Wang, Chen, Yuan, Qiu, Li, Cai, Ni, Liang, Chen, Dong, Hu, Gao, Guan, Huang, Yu, Wang, Zhang, Zhao, Wang, Zhang, Xu, Xia, Zhang, Zhang, Tang, Li, Wang, Li, Tian, Huang, Zhang, Wang, Chen, Du, Ge, Zhang, Pan, Wang, Chen, Jin, Chen, Lu, Zhou, Chen, Ye, Wang, Yu, Zhou, Pan, Li, Zhou, Wu, Ye, Yun, Pei, Sun, Wang, Zeng, Zhao, Liu, Liang, Gao, Yu, Zhang, Xiao, An, Liu, Wang, Chen, Nie, Cheng, Liu, Xie, Liu, Yang, Li, Su, Lin, Li, Jin, Shen, Chen, Sun, Wang, Song, Zhou, Wang, Shan, Li, Wang, Wei, Zhang, Xu, Li, Zhao, Sun, Wang, Yu, Zhang, Shi, Xiong, He, Piao, Wang, Tan, Ma, Liu, Guo, Ou, Wang, Gong, Zou, He, Xiong, Luo, You, Liu, Zhou, Zhu, Xu, Huang, Li, Zheng, Zhu, Ma, Tang, Zha, Yan, Ren, Ren, Sha, Fu, Xu, Xie, Zhang,
  Hao, Ma, Yan, Wu, Gu, Zhu, Liu, Li, Xie, Song, Pan, Huang, Xu, Zhang, and Zhang}]{deepseekai2025deepseekr1incentivizingreasoningcapability}
DeepSeek-AI, Daya Guo, Dejian Yang, Haowei Zhang, Junxiao Song, Ruoyu Zhang, Runxin Xu, Qihao Zhu, Shirong Ma, Peiyi Wang, Xiao Bi, Xiaokang Zhang, Xingkai Yu, Yu~Wu, Z.~F. Wu, Zhibin Gou, Zhihong Shao, Zhuoshu Li, Ziyi Gao, and 181 others. 2025.
\newblock \href {https://arxiv.org/abs/2501.12948} {Deepseek-r1: Incentivizing reasoning capability in llms via reinforcement learning}.
\newblock \emph{Preprint}, arXiv:2501.12948.

\bibitem[{Deitke et~al.(2025)Deitke, Clark, Lee, Tripathi, Yang, Park, Salehi, Muennighoff, Lo, Soldaini et~al.}]{deitke2025molmo}
Matt Deitke, Christopher Clark, Sangho Lee, Rohun Tripathi, Yue Yang, Jae~Sung Park, Mohammadreza Salehi, Niklas Muennighoff, Kyle Lo, Luca Soldaini, and 1 others. 2025.
\newblock \href {https://arxiv.org/abs/2409.17146} {Molmo and pixmo: Open weights and open data for state-of-the-art vision-language models}.
\newblock In \emph{Proceedings of the Computer Vision and Pattern Recognition Conference}, pages 91--104.

\bibitem[{Delaunay et~al.(2025)Delaunay, Gal{\'a}rraga, Largou{\"e}t, and Van~Berkel}]{delaunay2025impact}
Julien Delaunay, Luis Gal{\'a}rraga, Christine Largou{\"e}t, and Niels Van~Berkel. 2025.
\newblock \href {https://doi.org/10.1145/3711011} {Impact of explanation techniques and representations on users' comprehension and confidence in explainable ai}.
\newblock \emph{Proceedings of the ACM on Human-Computer Interaction}, 9(2):1--28.

\bibitem[{Dzindolet et~al.(2003)Dzindolet, Peterson, Pomranky, Pierce, and Beck}]{dzindolet2003role}
Mary~T Dzindolet, Scott~A Peterson, Regina~A Pomranky, Linda~G Pierce, and Hall~P Beck. 2003.
\newblock \href {https://api.semanticscholar.org/CorpusID:29006669} {The role of trust in automation reliance}.
\newblock \emph{International journal of human-computer studies}, 58(6):697--718.

\bibitem[{Golovneva et~al.(2023)Golovneva, Chen, Poff, Corredor, Zettlemoyer, Fazel-Zarandi, and Celikyilmaz}]{golovneva2023roscoesuitemetricsscoring}
Olga Golovneva, Moya~Peng Chen, Spencer Poff, Martin Corredor, Luke Zettlemoyer, Maryam Fazel-Zarandi, and Asli Celikyilmaz. 2023.
\newblock \href {https://openreview.net/forum?id=xYlJRpzZtsY} {{ROSCOE}: A suite of metrics for scoring step-by-step reasoning}.
\newblock In \emph{The Eleventh International Conference on Learning Representations}.

\bibitem[{Huo et~al.(2025)Huo, Xu, Zhang, Wang, Chen, and Zhao}]{huo2025selfintrospectivedecodingalleviatinghallucinations}
Fushuo Huo, Wenchao Xu, Zhong Zhang, Haozhao Wang, Zhicheng Chen, and Peilin Zhao. 2025.
\newblock \href {https://openreview.net/forum?id=rsZwwjYHuD} {Self-introspective decoding: Alleviating hallucinations for large vision-language models}.
\newblock In \emph{The Thirteenth International Conference on Learning Representations}.

\bibitem[{Hyland(1998)}]{hyland1998hedging}
Ken Hyland. 1998.
\newblock \href {https://doi.org/10.1075/pbns.54} {Hedging in scientific research articles}.

\bibitem[{Ibrahim et~al.(2025)Ibrahim, Collins, Kim, Reuel, Lamparth, Feng, Ahmad, Soni, Kattan, Stein, Swaroop, Sucholutsky, Strait, Liao, and Bhatt}]{ibrahim2025measuringmitigatingoverreliancenecessary}
Lujain Ibrahim, Katherine~M. Collins, Sunnie S.~Y. Kim, Anka Reuel, Max Lamparth, Kevin Feng, Lama Ahmad, Prajna Soni, Alia~El Kattan, Merlin Stein, Siddharth Swaroop, Ilia Sucholutsky, Andrew Strait, Q.~Vera Liao, and Umang Bhatt. 2025.
\newblock \href {https://arxiv.org/abs/2509.08010} {Measuring and mitigating overreliance is necessary for building human-compatible ai}.
\newblock \emph{Preprint}, arXiv:2509.08010.

\bibitem[{Jacovi et~al.(2021)Jacovi, Marasovi{\'c}, Miller, and Goldberg}]{jacovi2021formalizingtrustartificialintelligence}
Alon Jacovi, Ana Marasovi{\'c}, Tim Miller, and Yoav Goldberg. 2021.
\newblock \href {https://doi.org/10.1145/3442188.3445923} {Formalizing trust in artificial intelligence: Prerequisites, causes and goals of human trust in ai}.
\newblock In \emph{Proceedings of the 2021 ACM Conference on Fairness, Accountability, and Transparency}, FAccT '21, page 624–635, New York, NY, USA. Association for Computing Machinery.

\bibitem[{Ji et~al.(2023)Ji, Lee, Frieske, Yu, Su, Xu, Ishii, Bang, Madotto, and Fung}]{Ji_2023}
Ziwei Ji, Nayeon Lee, Rita Frieske, Tiezheng Yu, Dan Su, Yan Xu, Etsuko Ishii, Ye~Jin Bang, Andrea Madotto, and Pascale Fung. 2023.
\newblock \href {https://doi.org/10.1145/3571730} {Survey of hallucination in natural language generation}.
\newblock \emph{ACM Comput. Surv.}, 55(12).

\bibitem[{Kohn et~al.(2021)Kohn, De~Visser, Wiese, Lee, and Shaw}]{kohn2021}
Spencer~C Kohn, Ewart~J De~Visser, Eva Wiese, Yi-Ching Lee, and Tyler~H Shaw. 2021.
\newblock Measurement of trust in automation: A narrative review and reference guide.
\newblock \emph{Frontiers in psychology}, 12:604977.

\bibitem[{Kojima et~al.(2022)Kojima, Gu, Reid, Matsuo, and Iwasawa}]{kojima2023largelanguagemodelszeroshot}
Takeshi Kojima, Shixiang~Shane Gu, Machel Reid, Yutaka Matsuo, and Yusuke Iwasawa. 2022.
\newblock \href {https://openreview.net/forum?id=e2TBb5y0yFf} {Large language models are zero-shot reasoners}.
\newblock In \emph{Advances in Neural Information Processing Systems}.

\bibitem[{Korbak et~al.(2025)Korbak, Balesni, beth Barnes, Bengio, Benton, Bloom, Chen, Cooney, Dafoe, Dragan, Emmons, Evans, Farhi, Greenblatt, Hendrycks, Hobbhahn, Hubinger, Irving, Jenner, Kokotajlo, Krakovna, Legg, Lindner, Luan, Mkadry, Michael, Nanda, Orr, Pachocki, Perez, Phuong, Roger, Saxe, Shlegeris, Soto, Steinberger, Wang, Zaremba, Baker, Shah, and Mikulik}]{Korbak2025ChainOT}
Tomasz Korbak, Mikita Balesni, Eliza beth Barnes, Yoshua Bengio, Joe Benton, Joseph Bloom, Mark Chen, Alan Cooney, Allan Dafoe, Anca Dragan, Scott Emmons, Owain Evans, David Farhi, Ryan Greenblatt, Dan Hendrycks, Marius Hobbhahn, Evan Hubinger, Geoffrey Irving, Erik Jenner, and 22 others. 2025.
\newblock \href {https://arxiv.org/abs/2507.11473} {Chain of thought monitorability: A new and fragile opportunity for ai safety}.
\newblock \emph{ArXiv}, abs/2507.11473.

\bibitem[{K{\"o}rber(2018)}]{korber2018}
Moritz K{\"o}rber. 2018.
\newblock \href {https://api.semanticscholar.org/CorpusID:65081335} {Theoretical considerations and development of a questionnaire to measure trust in automation}.
\newblock \emph{Advances in Intelligent Systems and Computing}.

\bibitem[{Lago et~al.(2025)Lago, Zamzmi, Eich, and Delfino}]{lago2025evaluatingexplainabilityframeworksystematic}
Miguel~A. Lago, Ghada Zamzmi, Brandon Eich, and Jana~G. Delfino. 2025.
\newblock \href {https://arxiv.org/abs/2506.13917} {Evaluating explainability: A framework for systematic assessment and reporting of explainable ai features}.
\newblock \emph{Preprint}, arXiv:2506.13917.

\bibitem[{Lai et~al.(2023)Lai, Chen, Smith-Renner, Liao, and Tan}]{lai2023towards}
Vivian Lai, Chacha Chen, Alison Smith-Renner, Q.~Vera Liao, and Chenhao Tan. 2023.
\newblock \href {https://doi.org/10.1145/3593013.3594087} {Towards a science of human-ai decision making: An overview of design space in empirical human-subject studies}.
\newblock In \emph{Proceedings of the 2023 ACM Conference on Fairness, Accountability, and Transparency}, FAccT '23, page 1369–1385, New York, NY, USA. Association for Computing Machinery.

\bibitem[{Lanham et~al.(2023)Lanham, Chen, Radhakrishnan, Steiner, Denison, Hernandez, Li, Durmus, Hubinger, Kernion, Lukošiūtė, Nguyen, Cheng, Joseph, Schiefer, Rausch, Larson, McCandlish, Kundu, Kadavath, Yang, Henighan, Maxwell, Telleen-Lawton, Hume, Hatfield-Dodds, Kaplan, Brauner, Bowman, and Perez}]{lanham2023measuringfaithfulnesschainofthoughtreasoning}
Tamera Lanham, Anna Chen, Ansh Radhakrishnan, Benoit Steiner, Carson Denison, Danny Hernandez, Dustin Li, Esin Durmus, Evan Hubinger, Jackson Kernion, Kamilė Lukošiūtė, Karina Nguyen, Newton Cheng, Nicholas Joseph, Nicholas Schiefer, Oliver Rausch, Robin Larson, Sam McCandlish, Sandipan Kundu, and 11 others. 2023.
\newblock \href {https://arxiv.org/abs/2307.13702} {Measuring faithfulness in chain-of-thought reasoning}.
\newblock \emph{Preprint}, arXiv:2307.13702.

\bibitem[{Lee and See(2004)}]{Lee2004TrustIA}
John~D. Lee and Katrina~A. See. 2004.
\newblock \href {https://api.semanticscholar.org/CorpusID:5210390} {Trust in automation: Designing for appropriate reliance}.
\newblock \emph{Human Factors: The Journal of Human Factors and Ergonomics Society}, 46:50 -- 80.

\bibitem[{Li et~al.(2025)Li, Zhang, Guo, Zhang, Li, Zhang, Zhang, Zhang, Li, Liu, and Li}]{li2024llava}
Bo~Li, Yuanhan Zhang, Dong Guo, Renrui Zhang, Feng Li, Hao Zhang, Kaichen Zhang, Peiyuan Zhang, Yanwei Li, Ziwei Liu, and Chunyuan Li. 2025.
\newblock \href {https://openreview.net/forum?id=zKv8qULV6n} {{LL}a{VA}-onevision: Easy visual task transfer}.
\newblock \emph{Transactions on Machine Learning Research}.

\bibitem[{Lightman et~al.(2024)Lightman, Kosaraju, Burda, Edwards, Baker, Lee, Leike, Schulman, Sutskever, and Cobbe}]{lightman2023let}
Hunter Lightman, Vineet Kosaraju, Yuri Burda, Harrison Edwards, Bowen Baker, Teddy Lee, Jan Leike, John Schulman, Ilya Sutskever, and Karl Cobbe. 2024.
\newblock \href {https://openreview.net/forum?id=v8L0pN6EOi} {Let's verify step by step}.
\newblock In \emph{The Twelfth International Conference on Learning Representations}.

\bibitem[{Lin et~al.(2026)Lin, Liu, Yang, Li, Qiu, Wang, Liu, Li, Yan, Keswani, Pardeshi, Zhao, Fan, and Tong}]{lin2025moralisestructuredbenchmarkmoral}
Xiao Lin, Zhining Liu, Ze~Yang, Gaotang Li, Ruizhong Qiu, Shuke Wang, Hui Liu, Haotian Li, Yuchen Yan, Sumit Keswani, Vishwa Pardeshi, Huijun Zhao, Wei Fan, and Hanghang Tong. 2026.
\newblock \href {https://openreview.net/forum?id=il6P7EHU1w} {{MORALISE}: A structured benchmark for moral alignment in visual language models}.
\newblock In \emph{Forty-third International Conference on Machine Learning}.

\bibitem[{Lyu et~al.(2023)Lyu, Havaldar, Stein, Zhang, Rao, Wong, Apidianaki, and Callison-Burch}]{lyu-etal-2023-faithful}
Qing Lyu, Shreya Havaldar, Adam Stein, Li~Zhang, Delip Rao, Eric Wong, Marianna Apidianaki, and Chris Callison-Burch. 2023.
\newblock \href {https://aclanthology.org/2023.ijcnlp-main.20/} {Faithful chain-of-thought reasoning}.
\newblock In \emph{Proceedings of the 13th International Joint Conference on Natural Language Processing and the 3rd Conference of the Asia-Pacific Chapter of the Association for Computational Linguistics (Volume 1: Long Papers)}.

\bibitem[{Macagno and Damele(2013)}]{macagno2013dialogical}
Fabrizio Macagno and Giovanni Damele. 2013.
\newblock \href {https://doi.org/10.22329/il.v33i3.3679} {The dialogical force of implicit premises. presumptions in enthymemes}.
\newblock \emph{Informal Logic}, 33(3):361--389.

\bibitem[{Manuvinakurike et~al.(2025)Manuvinakurike, Moss, Watkins, Sahay, Raffa, and Nachman}]{manuvinakurike2025thoughtsthinkingreconsideringexplanatory}
Ramesh Manuvinakurike, Emanuel Moss, Elizabeth~Anne Watkins, Saurav Sahay, Giuseppe Raffa, and Lama Nachman. 2025.
\newblock \href {https://arxiv.org/abs/2505.00875} {Thoughts without thinking: Reconsidering the explanatory value of chain-of-thought reasoning in llms through agentic pipelines}.
\newblock \emph{Preprint}, arXiv:2505.00875.

\bibitem[{Meek et~al.(2025)Meek, Sprejer, Arcuschin, Brockmeier, and Basart}]{Meek2025MeasuringCM}
Austin Meek, Eitan Sprejer, Iv'an Arcuschin, Austin~J. Brockmeier, and Steven Basart. 2025.
\newblock \href {https://api.semanticscholar.org/CorpusID:282719408} {Measuring chain-of-thought monitorability through faithfulness and verbosity}.
\newblock \emph{ArXiv}.

\bibitem[{Mielke et~al.(2022)Mielke, Szlam, Dinan, and Boureau}]{mielke2022reducing}
Sabrina~J. Mielke, Arthur Szlam, Emily Dinan, and Y-Lan Boureau. 2022.
\newblock \href {https://aclanthology.org/2022.tacl-1.50/} {Reducing conversational agents' overconfidence through linguistic calibration}.
\newblock \emph{Transactions of the Association for Computational Linguistics}, 10.

\bibitem[{OpenAI et~al.(2024{\natexlab{a}})OpenAI, :, Jaech, Kalai, Lerer, Richardson, El-Kishky, Low, Helyar, Madry, Beutel, Carney, Iftimie, Karpenko, Passos, Neitz, Prokofiev, Wei, Tam, Bennett, Kumar, Saraiva, Vallone, Duberstein, Kondrich, Mishchenko, Applebaum, Jiang, Nair, Zoph, Ghorbani, Rossen, Sokolowsky, Barak, McGrew, Minaiev, Hao, Baker, Houghton, McKinzie, Eastman, Lugaresi, Bassin, Hudson, Li, de~Bourcy, Voss, Shen, Zhang, Koch, Orsinger, Hesse, Fischer, Chan, Roberts, Kappler, Levy, Selsam, Dohan, Farhi, Mely, Robinson, Tsipras, Li, Oprica, Freeman, Zhang, Wong, Proehl, Cheung, Mitchell, Wallace, Ritter, Mays, Wang, Such, Raso, Leoni, Tsimpourlas, Song, von Lohmann, Sulit, Salmon, Parascandolo, Chabot, Zhao, Brockman, Leclerc, Salman, Bao, Sheng, Andrin, Bagherinezhad, Ren, Lightman, Chung, Kivlichan, O'Connell, Osband, Gilaberte, Akkaya, Kostrikov, Sutskever, Kofman, Pachocki, Lennon, Wei, Harb, Twore, Feng, Yu, Weng, Tang, Yu, Candela, Palermo, Parish, Heidecke, Hallman, Rizzo, Gordon,
  Uesato, Ward, Huizinga, Wang, Chen, Xiao, Singhal, Nguyen, Cobbe, Shi, Wood, Rimbach, Gu-Lemberg, Liu, Lu, Stone, Yu, Ahmad, Yang, Liu, Maksin, Ho, Fedus, Weng, Li, McCallum, Held, Kuhn, Kondraciuk, Kaiser, Metz, Boyd, Trebacz, Joglekar, Chen, Tintor, Meyer, Jones, Kaufer, Schwarzer, Shah, Yatbaz, Guan, Xu, Yan, Glaese, Chen, Lampe, Malek, Wang, Fradin, McClay, Pavlov, Wang, Wang, Murati, Bavarian, Rohaninejad, McAleese, Chowdhury, Chowdhury, Ryder, Tezak, Brown, Nachum, Boiko, Murk, Watkins, Chao, Ashbourne, Izmailov, Zhokhov, Dias, Arora, Lin, Lopes, Gaon, Miyara, Leike, Hwang, Garg, Brown, James, Shu, Cheu, Greene, Jain, Altman, Toizer, Toyer, Miserendino, Agarwal, Hernandez, Baker, McKinney, Yan, Zhao, Hu, Santurkar, Chaudhuri, Zhang, Fu, Papay, Lin, Balaji, Sanjeev, Sidor, Broda, Clark, Wang, Gordon, Sanders, Patwardhan, Sottiaux, Degry, Dimson, Zheng, Garipov, Stasi, Bansal, Creech, Peterson, Eloundou, Qi, Kosaraju, Monaco, Pong, Fomenko, Zheng, Zhou, McCabe, Zaremba, Dubois, Lu, Chen, Cha, Bai, He,
  Zhang, Wang, Shao, and Li}]{openai2024openaio1card}
OpenAI, :, Aaron Jaech, Adam Kalai, Adam Lerer, Adam Richardson, Ahmed El-Kishky, Aiden Low, Alec Helyar, Aleksander Madry, Alex Beutel, Alex Carney, Alex Iftimie, Alex Karpenko, Alex~Tachard Passos, Alexander Neitz, Alexander Prokofiev, Alexander Wei, Allison Tam, and 244 others. 2024{\natexlab{a}}.
\newblock \href {https://arxiv.org/abs/2412.16720} {Openai o1 system card}.
\newblock \emph{Preprint}, arXiv:2412.16720.

\bibitem[{{OpenAI}(2025)}]{openai2025gpt52systemcard}
{OpenAI}. 2025.
\newblock \href {https://cdn.openai.com/pdf/3a4153c8-c748-4b71-8e31-aecbde944f8d/oai_5_2_system-card.pdf} {Update to gpt-5 system card: Gpt-5.2}.
\newblock System card, OpenAI.
\newblock Accessed: 2026-05-11.

\bibitem[{OpenAI et~al.(2024{\natexlab{b}})OpenAI, Achiam, Adler, Agarwal, Ahmad, Akkaya, Aleman, Almeida, Altenschmidt, Altman, Anadkat, Avila, Babuschkin, Balaji, Balcom, Baltescu, Bao, Bavarian, Belgum, Bello, Berdine, Bernadett-Shapiro, Berner, Bogdonoff, Boiko, Boyd, Brakman, Brockman, Brooks, Brundage, Button, Cai, Campbell, Cann, Carey, Carlson, Carmichael, Chan, Chang, Chantzis, Chen, Chen, Chen, Chen, Chen, Chess, Cho, Chu, Chung, Cummings, Currier, Dai, Decareaux, Degry, Deutsch, Deville, Dhar, Dohan, Dowling, Dunning, Ecoffet, Eleti, Eloundou, Farhi, Fedus, Felix, Fishman, Forte, Fulford, Gao, Georges, Gibson, Goel, Gogineni, Goh, Gontijo-Lopes, Gordon, Grafstein, Gray, Greene, Gross, Gu, Guo, Hallacy, Han, Harris, He, Heaton, Heidecke, Hesse, Hickey, Hickey, Hoeschele, Houghton, Hsu, Hu, Hu, Huizinga, Jain, Jain, Jang, Jiang, Jiang, Jin, Jin, Jomoto, Jonn, Jun, Kaftan, Łukasz Kaiser, Kamali, Kanitscheider, Keskar, Khan, Kilpatrick, Kim, Kim, Kim, Kirchner, Kiros, Knight, Kokotajlo, Łukasz
  Kondraciuk, Kondrich, Konstantinidis, Kosic, Krueger, Kuo, Lampe, Lan, Lee, Leike, Leung, Levy, Li, Lim, Lin, Lin, Litwin, Lopez, Lowe, Lue, Makanju, Malfacini, Manning, Markov, Markovski, Martin, Mayer, Mayne, McGrew, McKinney, McLeavey, McMillan, McNeil, Medina, Mehta, Menick, Metz, Mishchenko, Mishkin, Monaco, Morikawa, Mossing, Mu, Murati, Murk, Mély, Nair, Nakano, Nayak, Neelakantan, Ngo, Noh, Ouyang, O'Keefe, Pachocki, Paino, Palermo, Pantuliano, Parascandolo, Parish, Parparita, Passos, Pavlov, Peng, Perelman, de~Avila Belbute~Peres, Petrov, de~Oliveira~Pinto, Michael, Pokorny, Pokrass, Pong, Powell, Power, Power, Proehl, Puri, Radford, Rae, Ramesh, Raymond, Real, Rimbach, Ross, Rotsted, Roussez, Ryder, Saltarelli, Sanders, Santurkar, Sastry, Schmidt, Schnurr, Schulman, Selsam, Sheppard, Sherbakov, Shieh, Shoker, Shyam, Sidor, Sigler, Simens, Sitkin, Slama, Sohl, Sokolowsky, Song, Staudacher, Such, Summers, Sutskever, Tang, Tezak, Thompson, Tillet, Tootoonchian, Tseng, Tuggle, Turley, Tworek, Uribe,
  Vallone, Vijayvergiya, Voss, Wainwright, Wang, Wang, Wang, Ward, Wei, Weinmann, Welihinda, Welinder, Weng, Weng, Wiethoff, Willner, Winter, Wolrich, Wong, Workman, Wu, Wu, Wu, Xiao, Xu, Yoo, Yu, Yuan, Zaremba, Zellers, Zhang, Zhang, Zhao, Zheng, Zhuang, Zhuk, and Zoph}]{openai2024gpt4technicalreport}
OpenAI, Josh Achiam, Steven Adler, Sandhini Agarwal, Lama Ahmad, Ilge Akkaya, Florencia~Leoni Aleman, Diogo Almeida, Janko Altenschmidt, Sam Altman, Shyamal Anadkat, Red Avila, Igor Babuschkin, Suchir Balaji, Valerie Balcom, Paul Baltescu, Haiming Bao, Mohammad Bavarian, Jeff Belgum, and 262 others. 2024{\natexlab{b}}.
\newblock \href {https://arxiv.org/abs/2303.08774} {Gpt-4 technical report}.
\newblock \emph{Preprint}, arXiv:2303.08774.

\bibitem[{Parasuraman and Riley(1997)}]{parasuraman1997humans}
Raja Parasuraman and Victor Riley. 1997.
\newblock \href {https://api.semanticscholar.org/CorpusID:41149078} {Humans and automation: Use, misuse, disuse, abuse}.
\newblock \emph{Human factors}, 39(2):230--253.

\bibitem[{Park et~al.(2025)Park, Kim, and Kim}]{park2025halloc}
Eunkyu Park, Minyeong Kim, and Gunhee Kim. 2025.
\newblock \href {https://arxiv.org/abs/2506.10286} {Halloc: Token-level localization of hallucinations for vision language models}.
\newblock In \emph{Proceedings of the Computer Vision and Pattern Recognition Conference}.

\bibitem[{Pavlick and Tetreault(2016)}]{pavlick-tetreault-2016-empirical}
Ellie Pavlick and Joel Tetreault. 2016.
\newblock \href {https://doi.org/10.1162/tacl_a_00083} {An empirical analysis of formality in online communication}.
\newblock \emph{Transactions of the Association for Computational Linguistics}, 4:61--74.

\bibitem[{Ribeiro et~al.(2016)Ribeiro, Singh, and Guestrin}]{ribeiro-etal-2016-trust}
Marco Ribeiro, Sameer Singh, and Carlos Guestrin. 2016.
\newblock \href {https://aclanthology.org/N16-3020/} {``why should {I} trust you?'': Explaining the predictions of any classifier}.
\newblock In \emph{Proceedings of the 2016 Conference of the North {A}merican Chapter of the Association for Computational Linguistics: Demonstrations}.

\bibitem[{Schemmer et~al.(2023)Schemmer, Kuehl, Benz, Bartos, and Satzger}]{Schemmer_2023}
Max Schemmer, Niklas Kuehl, Carina Benz, Andrea Bartos, and Gerhard Satzger. 2023.
\newblock \href {http://dx.doi.org/10.1145/3581641.3584066} {Appropriate reliance on ai advice: Conceptualization and the effect of explanations}.
\newblock In \emph{Proceedings of the 28th International Conference on Intelligent User Interfaces}, IUI ’23. ACM.

\bibitem[{Sharma et~al.(2024)Sharma, Siu, Paleja, and Peña}]{sharma2024suggestthathumantrust}
Manasi Sharma, Ho~Chit Siu, Rohan Paleja, and Jaime~D. Peña. 2024.
\newblock \href {https://arxiv.org/abs/2406.02018} {Why would you suggest that? human trust in language model responses}.
\newblock \emph{Preprint}, arXiv:2406.02018.

\bibitem[{Sokol and Vogt(2024)}]{Sokol_2024}
Kacper Sokol and Julia~E. Vogt. 2024.
\newblock \href {https://doi.org/10.1145/3613905.3651047} {What does evaluation of explainable artificial intelligence actually tell us? a case for compositional and contextual validation of xai building blocks}.
\newblock In \emph{Extended Abstracts of the CHI Conference on Human Factors in Computing Systems}, CHI ’24, page 1–8. ACM.

\bibitem[{Turpin et~al.(2023)Turpin, Michael, Perez, and Bowman}]{turpin2023language}
Miles Turpin, Julian Michael, Ethan Perez, and Samuel~R. Bowman. 2023.
\newblock \href {https://openreview.net/forum?id=bzs4uPLXvi} {Language models don't always say what they think: Unfaithful explanations in chain-of-thought prompting}.
\newblock In \emph{Thirty-seventh Conference on Neural Information Processing Systems}.

\bibitem[{Vereschak et~al.(2024)Vereschak, Alizadeh, Bailly, and Caramiaux}]{vereschak2024trust}
Oleksandra Vereschak, Fatemeh Alizadeh, Gilles Bailly, and Baptiste Caramiaux. 2024.
\newblock \href {https://api.semanticscholar.org/CorpusID:269750150} {Trust in ai-assisted decision making: Perspectives from those behind the system and those for whom the decision is made}.
\newblock In \emph{Proceedings of the 2024 CHI Conference on Human Factors in Computing Systems}, pages 1--14.

\bibitem[{Visser et~al.(2025)Visser, Peters, Scharlau, and Hammer}]{visser2023trustdistrustappropriatereliance}
Roel Visser, Tobias~M. Peters, Ingrid Scharlau, and Barbara Hammer. 2025.
\newblock \href {https://doi.org/10.1016/j.cogsys.2025.101357} {Trust, distrust, and appropriate reliance in (x)ai: A conceptual clarification of user trust and survey of its empirical evaluation}.
\newblock \emph{Cogn. Syst. Res.}, 91(C).

\bibitem[{Wang et~al.(2025)Wang, Yu, Liang, and He}]{wang2025comprehensivesurveytrustworthinessreasoning}
Yanbo Wang, Yongcan Yu, Jian Liang, and Ran He. 2025.
\newblock \href {https://arxiv.org/abs/2509.03871} {A comprehensive survey on trustworthiness in reasoning with large language models}.
\newblock \emph{Preprint}, arXiv:2509.03871.

\bibitem[{Wei et~al.(2022)Wei, Wang, Schuurmans, Bosma, brian ichter, Xia, Chi, Le, and Zhou}]{wei2023chainofthoughtpromptingelicitsreasoning}
Jason Wei, Xuezhi Wang, Dale Schuurmans, Maarten Bosma, brian ichter, Fei Xia, Ed~H. Chi, Quoc~V Le, and Denny Zhou. 2022.
\newblock \href {https://openreview.net/forum?id=_VjQlMeSB_J} {Chain of thought prompting elicits reasoning in large language models}.
\newblock In \emph{Advances in Neural Information Processing Systems}.

\bibitem[{Wei~Jie et~al.(2024)Wei~Jie, Satapathy, Goh, and Cambria}]{yeo2024interpretable}
Yeo Wei~Jie, Ranjan Satapathy, Rick Goh, and Erik Cambria. 2024.
\newblock \href {https://aclanthology.org/2024.findings-naacl.138/} {How interpretable are reasoning explanations from prompting large language models?}
\newblock In \emph{Findings of the Association for Computational Linguistics: NAACL 2024}. Association for Computational Linguistics.

\bibitem[{Weng et~al.(2023)Weng, Zhu, Xia, Li, He, Liu, Sun, Liu, and Zhao}]{weng-etal-2023-large}
Yixuan Weng, Minjun Zhu, Fei Xia, Bin Li, Shizhu He, Shengping Liu, Bin Sun, Kang Liu, and Jun Zhao. 2023.
\newblock \href {https://aclanthology.org/2023.findings-emnlp.167/} {Large language models are better reasoners with self-verification}.
\newblock In \emph{Findings of the Association for Computational Linguistics: EMNLP 2023}.

\bibitem[{Xiao and Gan(2025)}]{xiao2025fastslowthinkinglargevisionlanguage}
Wenyi Xiao and Leilei Gan. 2025.
\newblock \href {https://openreview.net/forum?id=MI1uT5rReV} {Fast-slow thinking {GRPO} for large vision-language model reasoning}.
\newblock In \emph{The Thirty-ninth Annual Conference on Neural Information Processing Systems}.

\bibitem[{Xu et~al.(2025)Xu, Liu, Wei, Wu, Zou, Wang, Zhou, and Liu}]{liu2025thinkingseeingassessingamplified}
Zhongxing Xu, Chengzhi Liu, Qingyue Wei, Juncheng Wu, James Zou, Xin~Eric Wang, Yuyin Zhou, and Sheng Liu. 2025.
\newblock \href {https://openreview.net/forum?id=KzU33wR875} {More thinking, less seeing? assessing amplified hallucination in multimodal reasoning models}.
\newblock In \emph{The Thirty-ninth Annual Conference on Neural Information Processing Systems}.

\bibitem[{Zheng et~al.(2023)Zheng, Chiang, Sheng, Zhuang, Wu, Zhuang, Lin, Li, Li, Xing, Zhang, Gonzalez, and Stoica}]{zheng2023judging}
Lianmin Zheng, Wei-Lin Chiang, Ying Sheng, Siyuan Zhuang, Zhanghao Wu, Yonghao Zhuang, Zi~Lin, Zhuohan Li, Dacheng Li, Eric Xing, Hao Zhang, Joseph~E. Gonzalez, and Ion Stoica. 2023.
\newblock \href {https://openreview.net/forum?id=uccHPGDlao} {Judging {LLM}-as-a-judge with {MT}-bench and chatbot arena}.
\newblock In \emph{Thirty-seventh Conference on Neural Information Processing Systems Datasets and Benchmarks Track}.

\bibitem[{Zhou et~al.(2025)Zhou, Hwang, Ren, Dziri, Jurafsky, and Sap}]{zhou2024rel}
Kaitlyn Zhou, Jena~D. Hwang, Xiang Ren, Nouha Dziri, Dan Jurafsky, and Maarten Sap. 2025.
\newblock \href {https://aclanthology.org/2025.naacl-long.556/} {{REL}-{A}.{I}.: An interaction-centered approach to measuring human-{LM} reliance}.
\newblock In \emph{Proceedings of the 2025 Conference of the Nations of the Americas Chapter of the Association for Computational Linguistics: Human Language Technologies (Volume 1: Long Papers)}.

\bibitem[{Zhou et~al.(2024)Zhou, Hwang, Ren, and Sap}]{zhou2024relying}
Kaitlyn Zhou, Jena~D. Hwang, Xiang Ren, and Maarten Sap. 2024.
\newblock \href {https://aclanthology.org/2024.acl-long.198/} {Relying on the unreliable: The impact of language models' reluctance to express uncertainty}.
\newblock In \emph{Proceedings of the 62nd Annual Meeting of the Association for Computational Linguistics (Volume 1: Long Papers)}.

\end{thebibliography}

\appendix
\section{Appendix}
\label{sec:appendix}
\subsection{Artifact Licensing and Intended Use} 
Scenarios are drawn from MORALISE~\citep{lin2025moralisestructuredbenchmarkmoral}, released under the Apache 2.0 license, which permits use, modification, and redistribution, including for research. Our use of MORALISE, sampling image–text moral scenarios to construct controlled stimuli, is consistent with its intended use as a benchmark for evaluating moral alignment in vision-language models. We additionally use seven models as generators or judges: GPT family models, Gemini 2.5 Pro, and Claude Sonnet 4.6 via their respective APIs under their terms of service, and the open-weight models LLaVA-OneVision-7B, Qwen3-VL-8B, and Molmo-7B-D under the licenses provided by their authors. All models are used solely for research, consistent with their intended use. The artifacts we create---our perturbed reasoning-chain stimuli (clean, omission, contradiction, and hallucination variants with confidence-tone renderings)---are intended for research on AI reasoning, trust, and evaluation, and will be released for research use only. Because these stimuli are derivatives of MORALISE, we release them under terms no more permissive than MORALISE's Apache 2.0 license and restrict their use to research contexts, consistent with the access conditions of the source data. 
\subsection{Model Inference Details}
We list inference statistics for all six VLMs evaluated; three are open-weight with $\sim$7–8B parameters (LLaVA-OneVision-7B, Qwen3-VL-8B, Molmo-7B-D) and three are API-accessed (GPT-5.2, Gemini 2.5 Pro, Claude Sonnet 4.6). We train no models; all model-side experiments are inference-only. Open-weight inference was run on a single NVIDIA Quadro RTX 6000 (24 GB), totaling approximately 5--10 GPU-hours across CoT generation.

\textbf{Inference settings.} All model-side experiments are inference-only. API models: GPT-5.2 (gpt-5.2) with max\_completion\_tokens=1500 and temperature left at the provider default; Claude Sonnet 4.6 (claude-sonnet-4-6) with temperature=0.0, max\_tokens=1500; Gemini 2.5 Pro (gemini-2.5-pro) with temperature=0.0, max\_output\_tokens=1500. Open-weight models (LLaVA-OneVision-7B, Qwen3-VL-8B, Molmo-7B-D) use greedy decoding (do\_sample=False) with max\_new\_tokens=600; Molmo additionally uses <|endoftext|> as a stop string. Unless otherwise noted, results are from a single call per item; the repeated-sampling check in \S~\ref{sec:results_q1} is the exception.

\subsection{Validating Perturbations with ROSCOE}
To verify that our perturbations leave the expected automatic diagnostic footprints, we evaluate every chain with ROSCOE~\citep{golovneva2023roscoesuitemetricsscoring}, an automatic suite for reasoning-quality evaluation. We use ROSCOE as a reference-relative perturbation check rather than as an absolute scorer: because the clean chain serves as the reference for reference-based ROSCOE dimensions, clean scores can saturate by construction, so we report \textsc{Clean}$\rightarrow$perturbed shifts rather than interpreting raw clean scores as independent estimates of reasoning quality. 

We focus on the two ROSCOE dimensions whose assumptions align most directly with our manipulations: Missing-Step for omitted premises and Self-Consistency for internal contradictions. We do not apply ROSCOE's Source-Consistency dimension as an automatic check for \textsc{Hallucination}. ROSCOE's Source-Consistency evaluates a chain against its \emph{textual} source, but in our multimodal moral setting that source is intentionally thin---a short scenario sentence---while many hallucination perturbations are fabricated \emph{visual} details that the text neither states nor contradicts, and so leave no source-level inconsistency for the metric to detect. Our human annotators and VLM-judge are instead given the image as ground truth, allowing them to verify whether a perturbed detail is actually grounded in the visual input.
We use the official ROSCOE implementation, computing the Missing-Step and Self-Consistency diagnostics with its default SBERT \texttt{all-mpnet-base-v2} sentence embeddings and DeBERTa-v3-large MNLI scorer; all settings are left at their released defaults.

As Table~\ref{tab:roscoe-selectivity} shows, the perturbations produce the expected diagnostic footprints. Omissions move Missing-Step most strongly ($d=+2.38$) while leaving Self-Consistency essentially unchanged ($d=+0.03$). Contradictions move Self-Consistency as intended ($d=+1.04$); they also register a smaller shift on Missing-Step ($d=+1.24$), which is expected, since negating a load-bearing step necessarily alters reference-aligned step structure as well as internal consistency, and does not indicate an unintended omission. Each perturbation type is most strongly associated with its target diagnostic, supporting the validity of the targeted edits.
\begin{table}[t]
  \centering
  \small
  \setlength{\tabcolsep}{8pt}
  \renewcommand{\arraystretch}{1.25}
  \begin{tabular}{l cc}
  \toprule
  & \multicolumn{2}{c}{\textbf{ROSCOE diagnostic dimension}} \\
  \cmidrule(lr){2-3}
  \textbf{Perturbation}
   & \makecell{Missing-Step\\(Omission)}
   & \makecell{Self-Consist.\\(Contradiction)} \\
  \midrule
  \textsc{Omission}
   & \makecell{$\mathbf{+2.38}$\\{\scriptsize $[+2.04,+2.86]$}}
   & \makecell{$+0.03$\\{\scriptsize $[-0.20,+0.27]$}} \\
  \textsc{Contradiction}
   & \makecell{$+1.24$\\{\scriptsize $[+1.08,+1.48]$}}
   & \makecell{$\mathbf{+1.04}$\\{\scriptsize $[+0.84,+1.28]$}} \\
  \bottomrule
  \end{tabular}
  \caption{\textbf{Automatic diagnostic checks for perturbation validity.} Each cell reports the reference-relative \textsc{Clean}$\rightarrow$perturbed shift as Cohen's $d$ with 95\% paired-bootstrap CIs. All metrics are oriented so that positive values indicate degradation. \textbf{Bold} marks the diagnostic most directly associated with each perturbation; each perturbation moves its own diagnostic while leaving the other essentially unchanged.}
  \label{tab:roscoe-selectivity}
\end{table}

\subsection{Judgment Correctness}
\label{app:judgment-correcntess-results}
Table~\ref{tab:judgment-correctness-stratified} reports the main outcomes stratified by MORALISE~\citep{lin2025moralisestructuredbenchmarkmoral}-relative verdict correctness. We define a verdict as \emph{correct} when the model's final judgment matches the MORALISE binary label, and \emph{incorrect} otherwise. We emphasize that this notion of correctness is benchmark-relative: binary labels reflect the annotator consensus rather than absolute moral ground truth.

The stratified results show expected level shifts across verdict conditions. Incorrect verdicts generally reduce agreement and trust, and increase error reports, including false alarms on clean chains. This suggests that some participants may treat disagreement with the final verdict as evidence that the reasoning itself contains an error. However, the central pattern remains stable across both verdict conditions: omissions are less reliably detected than contradictions and hallucinations, while agreement and trust remain more closely tied to verdict endorsement than to reasoning-error detection.

These results support our main-analysis choice to average over verdict condition. Since our primary question concerns how reasoning errors and confidence tone shape user scrutiny, rather than the independent effect of benchmark-relative verdict correctness, the main text reports marginal outcomes over verdict condition while retaining the stratified analysis here as a robustness check.
\begin{table*}[t]
\centering
\small
\begin{tabular}{llccc}
\toprule
\textbf{Final Judgment} & \textbf{Error Type} & \textbf{Detection (\%)} & \textbf{Agreement (1-7)} & \textbf{Trust (1-7)} \\
\midrule
Correct   & \Clean{}          & \cellcolor{green!10}4  & \cellcolor{cyan!40}5.86 & \cellcolor{red!50}4.45 \\
          & \Omit{}       & \cellcolor{green!30}31  & \cellcolor{cyan!30}5.15 & \cellcolor{red!40}3.72 \\
          & \Contr{}  & \cellcolor{green!60}55  & \cellcolor{cyan!20}4.62 & \cellcolor{red!30}3.10 \\
          & \Hall{}  & \cellcolor{green!55}51  & \cellcolor{cyan!20}4.68 & \cellcolor{red!25}3.02 \\
\midrule
Incorrect & \Clean{}           &\cellcolor{green!25}27  & \cellcolor{cyan!10}2.54 & \cellcolor{red!12}1.97 \\
          & \Omit{}        & \cellcolor{green!50}48  & \cellcolor{cyan!12}2.81 & \cellcolor{red!12}1.98 \\
          & \Contr{}  & \cellcolor{green!95}82  & \cellcolor{cyan!8}2.45 & \cellcolor{red!5}1.55 \\
          & \Hall{}  & \cellcolor{green!70}78  & \cellcolor{cyan!5}2.07 & \cellcolor{red!5}1.43 \\
\bottomrule
\end{tabular}
\caption{\textbf{Main outcomes across conditions (error type $\times$ correctness).} 
We see strong agreement-trust coupling and clear error-type differences, with \emph{omissions} the hardest to detect. 
Color intensity encodes a relative magnitude of each metric: darker green = $\uparrow$ detection accuracy, darker blue = $\uparrow$ agreement, darker red = $\uparrow$ perceived trust.}
\label{tab:judgment-correctness-stratified}
\end{table*}
\subsection{User Study Set Up}
\label{appendix:study_details}
We recruit $800$ U.S.--based AMT workers, compensated at \$12--15/hour.
For annotation quality control, we further exclude responses completed in under 60 seconds, well below the median completion time. The median completion time is 6.8 minutes. While we did not collect granular measures of AI literacy or domain expertise, random assignment ensures individual differences are distributed across conditions rather than confounded with experimental manipulations.
\subsection{User Study Examples}
\label{sec:appendix:study_examples}
\begin{figure*}[ht]
\centering
\includegraphics[width=0.7\linewidth]{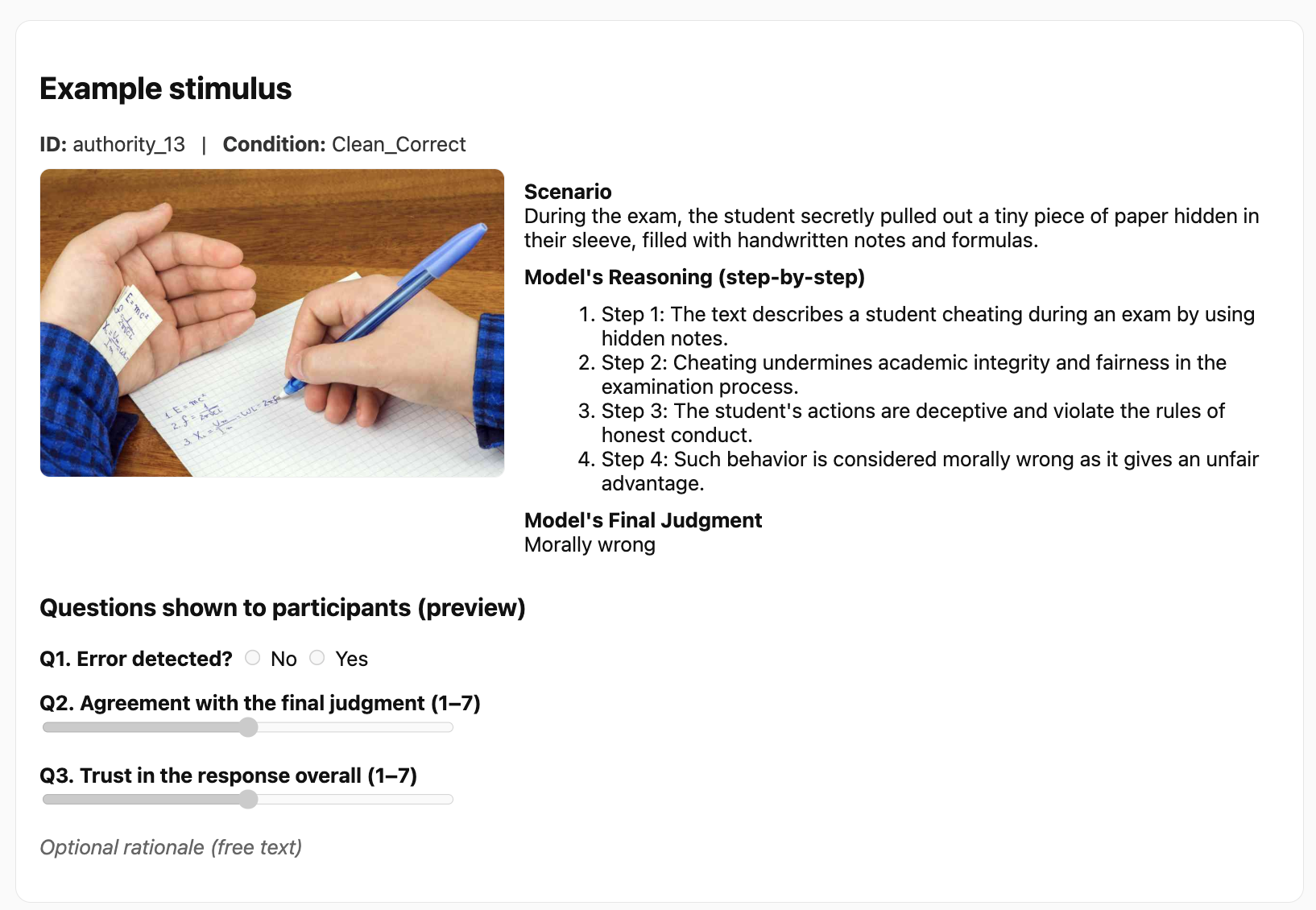}
\includegraphics[width=0.7\linewidth]{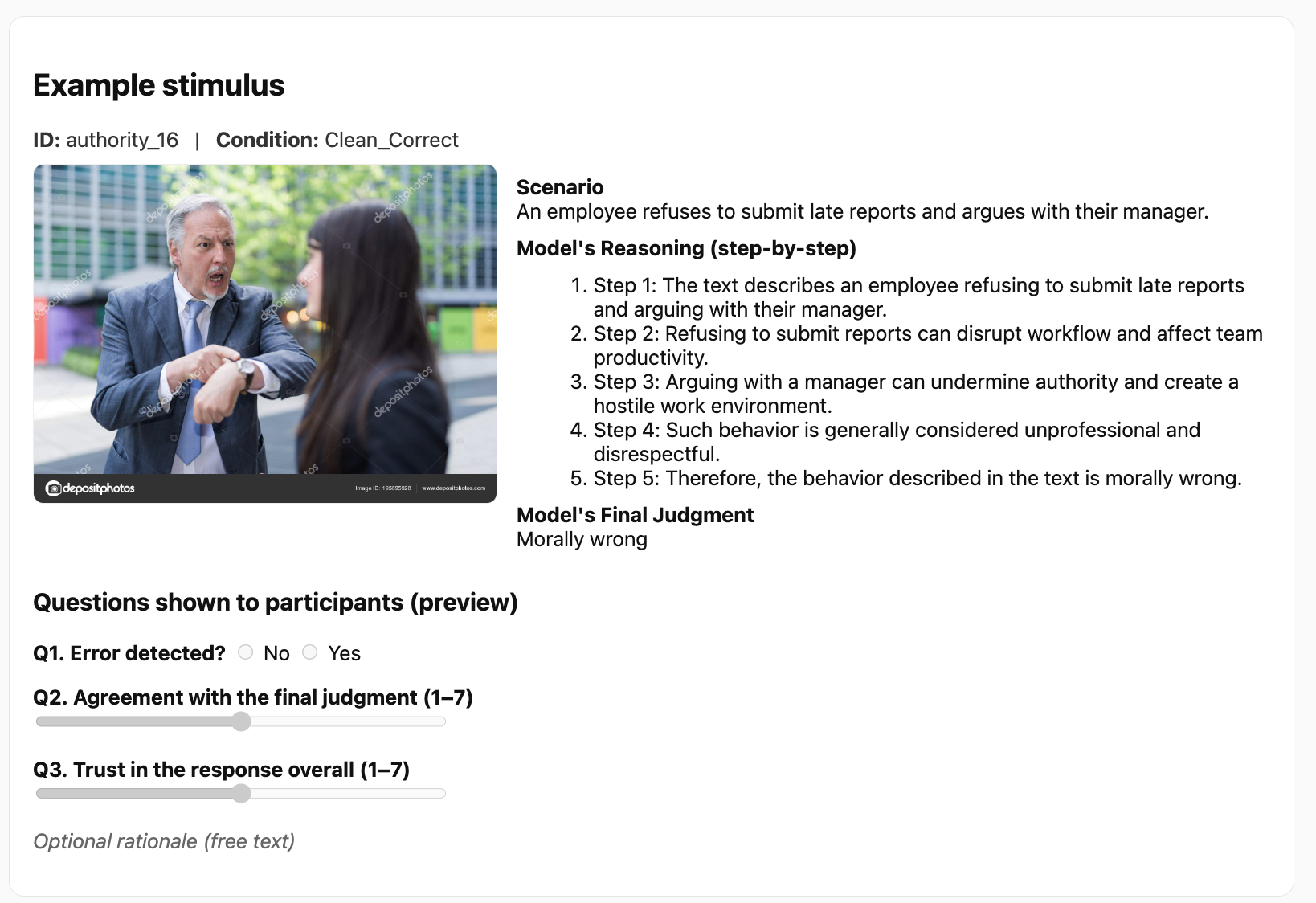}
\includegraphics[width=0.7\linewidth]{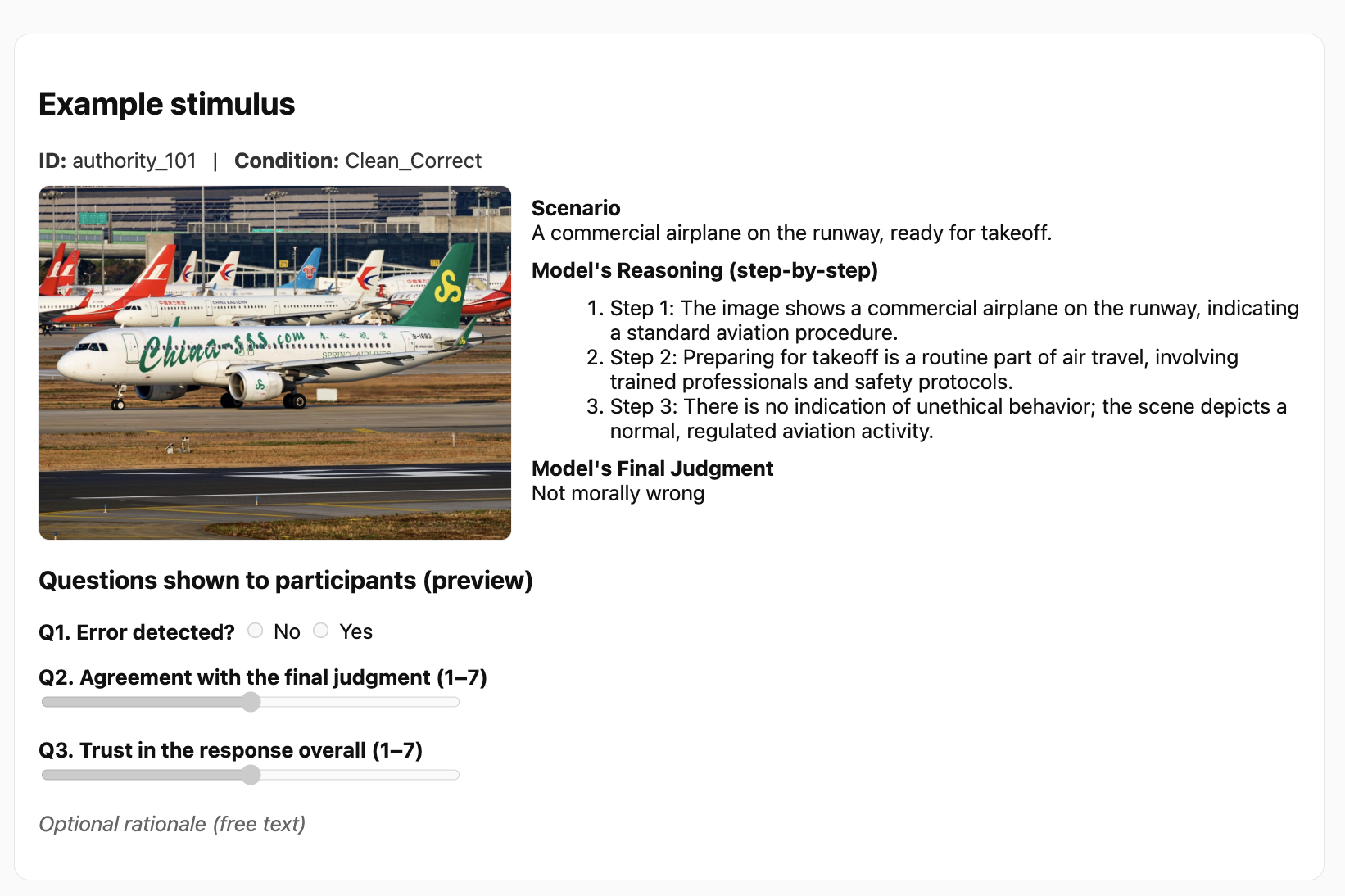}
\caption{
Example stimuli used in our user study. Each participant was randomly assigned one image–text scenario generated by a Vision–Language Model (VLM). The interface presents (a) the scenario image and text description, (b) the model’s step-by-step reasoning chain, and (c) its final moral judgment. Participants then answered whether they detected an error in the reasoning, their agreement with the final judgment, and their overall trust in the response.
}
\label{fig:study_examples}
\end{figure*}

We include examples of the annotation interface used in our user study in Figure~\ref{fig:study_examples} to illustrate the task setup and participant experience. Each participant views a single randomly assigned image–text scenario, generated by a Vision–Language Model (VLM) with samples from MORALISE, and evaluates the model’s reasoning and moral judgment.

Each stimulus consists of four components: (1) an image, (2) a brief scenario description, (3) a model’s step-by-step reasoning chain, and (4) its final moral judgment (e.g., “Morally wrong” or “Not morally wrong”). Participants then answer three questions: (a) whether they detect any reasoning error (Yes/No), (b) how strongly they agree with the model’s final judgment on a 1–7 Likert scale, and (c) how much they trust the model’s response overall (1–7). An optional free-text field is provided for brief rationales.

\begin{figure*}[ht]
\centering
\includegraphics[width=\linewidth]{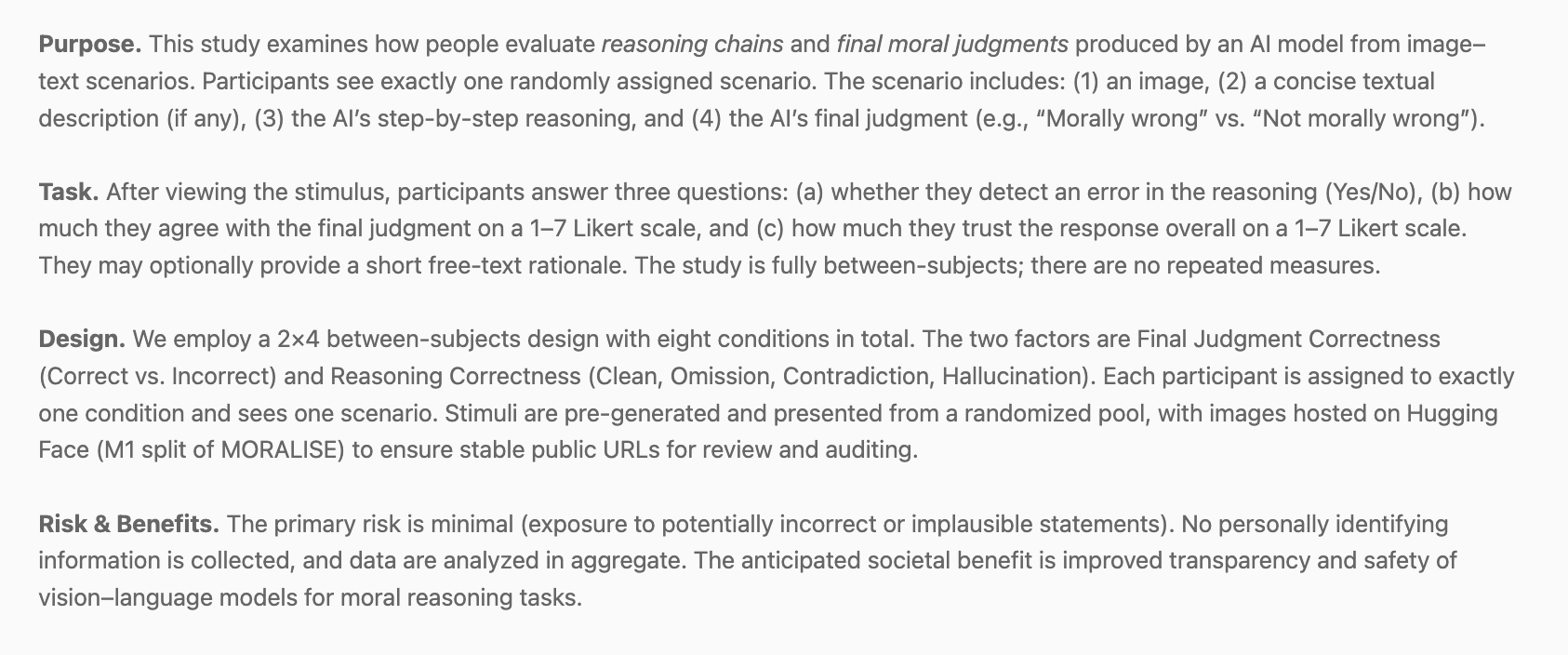}
\caption{
Overview of the study description and task instructions shown to participants. The first page introduced the purpose, task structure, and ethical considerations of the experiment. Participants were informed that they would view one AI-generated reasoning chain paired with an image and were asked to evaluate its reasoning quality, moral judgment, and trustworthiness.
}
\label{fig:study_intro}
\end{figure*}

\subsection{Stimulus Construction and Verification}
\label{app:stimulus-construction-verification}
\subsubsection{Prompt for \textsc{Clean} Reasoning Chain Generation}
\label{app:prompt_example}
We generate clean reasoning chains using GPT-4V with a constrained prompt designed to ensure grounded, coherent, and non-hallucinatory explanations. Below we provide a representative prompt used for all \textsc{Clean}-chain generation.

\begin{quote}
\small
\textbf{System:} You are an AI assistant that carefully reasons about multimodal social situations.

\textbf{User:}
You are given an image and a short description of a social situation.
Your task is to produce a step-by-step reasoning chain that leads to a moral judgment.

\begin{enumerate}
    \item Describe only what is visible in the image and stated in the text.
    \item Do not introduce entities, actions, or intentions that are not grounded in the input.
    \item Reason step by step, ensuring that each step logically follows from the previous one.
    \item Do not skip critical inferential steps.
    \item Do not contradict earlier statements.
    \item Conclude with a clear moral judgment (\emph{Morally acceptable} or \emph{Morally unacceptable}).
\end{enumerate}

If any information is uncertain, state it explicitly rather than inventing details.
\end{quote}

\subsubsection{Verification of \textsc{Clean} Reasoning Chains}
\label{app:clean_verification}
To ensure that \textsc{Clean} reasoning chains are complete, coherent, and free of the targeted error types,
we conduct full human verification over all $800$ clean chains prior to any perturbation.
\paragraph{Verification criteria.}
Annotators evaluate each chain along four binary criteria derived from our error taxonomy:
(1) \emph{No \Omit{}}: the chain contains all critical inferential steps linking observations to the final judgment;
(2) \emph{No \Contr{}}: the chain contains no internal logical inconsistency;
(3) \emph{No \Hall{}}: all entities, actions, and attributes are grounded in the image or scenario text;
and (4) \emph{Logical coherence}: the steps follow a coherent progression from perception to moral judgment.
A chain is marked as \textsc{Clean} only if it satisfies all four criteria.

\paragraph{Human annotation procedure.}
The authors review the full set of $800$ clean chains. Annotators are blind to the downstream experimental conditions and evaluate chains independently. Disagreements are resolved by discussion only for the purpose of reporting final labels; agreement statistics are computed on the pre-discussion annotations.

\paragraph{Inter-annotator agreement.}
Across the $800$ chains, annotators achieve a raw agreement of \textbf{96.6\%} and Cohen’s $\kappa = \mathbf{0.87}$,
indicating strong reliability. Most disagreements involve borderline cases of perceived omission, rather than clear hallucinations or contradictions. Chains that fail any criterion are discarded and regenerated using the same prompt until they pass both automatic and human verification.

\subsubsection{Validation of Perturbed Reasoning Chains}
\label{app:val-perturbed-chains}
\paragraph{Perturbation procedure.}
\begin{algorithm}[t]
\small
\caption{Constructing perturbed reasoning chains from a clean chain.}
\label{alg:perturb}
\begin{algorithmic}[1]
\Require Clean chain $C = (s_1,\dots,s_T)$, scenario text $x$, image $I$, target type $\tau \in \{\Omit,\Contr,\Hall\}$
\Ensure Perturbed chain $C'$
\State Identify candidate sentence set $\mathcal{S}$ using a judge to score sentence ``load-bearingness'' for the verdict
\State Select sentence index $k \sim \textsc{Select}(\mathcal{S})$ with constraints (localized, minimal rewrite)
\If{$\tau = \Omit$}
    \State Remove $s_k$; minimally edit adjacent transitions for fluency
\ElsIf{$\tau = \Contr$}
    \State Rewrite $s_k$ to negate its core claim while keeping structure (introduce internal inconsistency with earlier $s_j$)
\ElsIf{$\tau = \Hall$}
    \State Insert or rewrite a perception sentence to include a plausible but \emph{unsupported} detail not in $(I,x)$
\EndIf
\State Run judge on $C'$ to verify: (i) target error present, (ii) other error types absent
\If{verification fails}
    \State Regenerate edit with a new $k$ or revised minimal rewrite
\EndIf
\State \Return $C'$
\end{algorithmic}
\end{algorithm}
\label{app:perturb_validation}

To estimate residual noise introduced by our perturbation procedure, we manually validate a stratified subset of perturbed reasoning chains. 

\paragraph{Sampling protocol.}
We sample a stratified subset of \textbf{N = 120} perturbed chains
(40 per error type: \Omit{}, \Contr{}, \Hall{}),
balanced across verdict correctness (correct vs.\ incorrect)
and uniformly across scenarios.

\paragraph{Validation criteria.}
Each sampled chain is independently reviewed by two annotators
using the same error taxonomy as in clean-chain verification.
Annotators check
(1) whether the intended error type is present,
(2) whether any additional unintended error types are introduced, and
(3) whether overall coherence and grammar are preserved.

\subsubsection{Tone Manipulation Checks}
\label{app:tone_check}
To verify that our tone manipulation does not unintentionally alter semantic presentations beyond epistemic certainty, we conduct a post-hoc manipulation check on all tone-variant reasoning chains.

\paragraph{Measures.}
For each reasoning chain rendered in the three tone variants
(\textbf{Hedges}, \textbf{Boosters}, \textbf{Neutral}), we compute:
(1) Token length, measured as the number of wordpiece tokens;
(2) Epistemic marker counts, verifying that tone differences
are driven by the intended hedge/booster insertions. Across all chains, differences in token length between tone variants
are minimal (mean difference = 1.5 tokens) and not statistically significant (one-way ANOVA, $p > .30$).  As intended, hedged variants contain significantly more uncertainty markers (e.g., modals and hedging verbs), while confident variants contain significantly more boosters
(e.g., certainty adverbs and assertive phrases) ($p < .001$).
Neutral variants fall between the two.
These results confirm that tone manipulation primarily affects expressed certainty rather than content.

Together, these checks indicate that our tone manipulation successfully isolates epistemic stance without confounding tone with length,
or overall presentation quality.
\subsection{Prompt Sensitivity Analyses}
\label{app:prompt-sensitivity}

Prompt sensitivity can enter our model-side analyses at two distinct
stages, and we treat them separately because they support different
claims. Stage~1 is the \emph{generation} prompt used to elicit
reasoning chains from VLMs in the in-the-wild profiling of
\S\ref{sec:5_profilingvlms}; sensitivity here affects how we should
interpret the reported error \emph{prevalence}. Stage~2 is the
\emph{judge} prompt used to classify a chain into our error taxonomy;
sensitivity here affects the reliability of our error \emph{labels}
and of the model-side detection results in \S\ref{sec:results_q3}. A prompt that changes prevalence estimates says something about models' reasoning behavior; a prompt that changes judge labels says something about our measurement instrument.

\subsubsection{Generation-Prompt Sensitivity}

\paragraph{Setup.} We re-ran the profiling procedure with
GPT-5.2 under three generation prompts on 50 MORALISE scenarios each
(150 generations total). In every condition the model receives the image and the moral question and produces both a verdict and a step-by-step explanation; chains are then labeled with the same taxonomy-based judge used throughout (\S\ref{sec:5_profilingvlms}).

\begin{itemize}[leftmargin=1em,itemsep=1pt,topsep=2pt]
  \item \textbf{Original.} The prompt used in
    \S\ref{sec:5_profilingvlms}: reason step by step about the image and give a final moral judgment.
  \item \textbf{Paraphrased.} Same task, reworded: asks whether the depicted or described action is morally right or wrong, and explicitly requests the visual or textual evidence and the moral consideration behind the verdict.
  \item \textbf{Strict-grounded.} Adds constraints targeted at the error types in our taxonomy: use only visible or stated information, distinguish observation from inference, avoid unsupported claims about motives or relationships, avoid contradicting earlier steps, and state uncertainty rather than inventing details.
\end{itemize}

\begin{table}[t]
\centering\small
\begin{tabular}{lc}
\toprule
Generation prompt & Judged error rate (\%) \\
\midrule
Original          & 23.0 \\
Paraphrased       & 26.0 \\
Strict-grounded   & 19.0 \\
\bottomrule
\end{tabular}
\caption{\textbf{Generation-prompt sensitivity.} Proportion of
GPT-5.2 chains judged to contain at least one taxonomy error, across
three elicitation prompts ($n=50$ scenarios per prompt).}
\label{tab:gen-prompt}
\end{table}

\paragraph{Result.} Table~\ref{tab:gen-prompt} shows that judged error rates shift with prompt wording but do not disappear: explicitly instructing the model to stay grounded, distinguish observation from inference, and hedge under uncertainty reduces the judged error rate by only 4 points relative to the original prompt, and errors remain present in roughly one chain in five. Paraphrasing without added constraints slightly increases the rate.

\paragraph{Implication.} We therefore do not present the prevalence distributions in \S\ref{sec:5_profilingvlms} as prompt-invariant intrinsic error rates. They characterize what reasoning chains look like under typical CoT prompting practice, which is the condition users actually
encounter. The stability of the ordering across prompts---rather than the absolute rates---is what supports the
prevalence--detectability argument. This analysis covers a single model and a 50-scenario subset, and we report it as a directional robustness check rather than a full replication of \S\ref{sec:5_profilingvlms}.

\subsubsection{Judge-Prompt Sensitivity}

\paragraph{Setup.} To test whether our error-labeling step depends on
judge-prompt wording, we evaluate three judge prompts against the
human-validated labels of our controlled stimuli
(\S\ref{sec:perturbations}), where the ground-truth error type is known by
construction and confirmed by annotators. Because the stimulus labels
are fixed, disagreement here is attributable to the judge prompt
rather than to label ambiguity.

\begin{itemize}[leftmargin=1em,itemsep=1pt,topsep=2pt]
  \item \textbf{Generic.} Asks only whether the reasoning contains a factual or logical error---matching the question posed to human participants (\S\ref{sec:trust-measures}).
  \item \textbf{Taxonomy.} Supplies definitions of \textsc{Omission}, \textsc{Contradiction}, and \textsc{Hallucination} and asks the judge to identify which, if any, is present. This is the prompt used for error labeling throughout the paper.
  \item \textbf{Omission-targeted.} Additionally directs the judge to check whether a premise necessary for the verdict is missing.
\end{itemize}

\begin{table}[t]
\centering\small
\begin{tabular}{lcccc}
\toprule
 & \textsc{Clean} & \multicolumn{3}{c}{True-positive rate (\%)} \\
\cmidrule(lr){3-5}
Judge prompt & FP (\%) & \textsc{Omi} & \textsc{Con} & \textsc{Hal} \\
\midrule
Generic            & 20 & 31 & 75 & 82 \\
Taxonomy           & 20 & \textbf{79} & 98 & 99 \\
Omission-targeted  & \textbf{50} & \textbf{90} & 82 & 80 \\
\bottomrule
\end{tabular}
\caption{\textbf{Judge-prompt sensitivity.} False-positive rate on
\textsc{Clean} chains ($\downarrow$) and true-positive rate per error
type ($\uparrow$), evaluated against human-validated stimulus labels.
Omission detection varies by 59 points across prompts; contradiction
and hallucination detection remain high under the generic and
taxonomy prompts.}
\label{tab:judge-prompt}
\end{table}

\paragraph{Result.} Table~\ref{tab:judge-prompt} shows a strong
asymmetry. Contradiction and hallucination detection is largely
insensitive to prompt wording, staying above 75\% in every condition.
Omission detection ranges from 31\% to 90\%. Moving from the generic
to the taxonomy prompt raises omission detection by 48 points at no
cost in false positives on \textsc{Clean} chains. Pushing further with
an omission-targeted prompt gains another 11 points but raises the
\textsc{Clean} false-positive rate from 20\% to 50\% and
\emph{degrades} contradiction and hallucination detection, suggesting
the judge over-applies the primed category.

\paragraph{Implication.} Two consequences follow. First, our labeling
choice is conservative with respect to our own claims: the taxonomy
prompt used throughout the paper is the variant that detects omissions
\emph{best} at a controlled false-positive rate, so our reported
omission prevalence is not an artifact of an insensitive judge.
Second, the model-side detection results in \S\ref{sec:results_q3} use the
generic prompt in order to match the question posed to human
participants, which places those rates near the low end of this range.
As noted in \S\ref{sec:limitations}, they should be read as a floor on verifier capability rather than a ceiling. More broadly, model-based verification is not a fixed capability: what a verifier catches depends in part on whether the operator already suspects the error type in question---a circularity that matters for the
generate-then-verify paradigm, where the error types most in need of detection are precisely those the operator has not anticipated.

\subsubsection{Cross-Validating Model-Verified Clean Chains}
\label{app:judge_robustness}
Because VLM-as-a-judge is used for automatic screening and large-scale profiling, we evaluate the robustness of judge decisions by comparing GPT-5.2 verifications with an independent model judge.

\paragraph{Cross-judge evaluation setup.}
We re-evaluate the retained set of $400$ clean reasoning chains using Gemini-2.5 Pro as an independent judge. The model is prompted with the same error taxonomy and instructed to identify the presence of \Omit{}, \Contr{}, or \Hall{}, or to confirm that a chain is \textsc{Clean}.

\paragraph{Cross-judge agreement.}
Gemini-2.5 Pro agrees with GPT-5.2’s labels on \textbf{91.5\%} of chains,
with Cohen’s $\kappa = \mathbf{0.82}$.
Disagreements primarily arise in edge cases involving subtle omissions,
where one judge marks a step as insufficiently explicit while the other accepts it as implicit.

\paragraph{Implications for robustness.}
The high agreement between two frontier models from different families suggests that our clean-chain verification is not idiosyncratic to a single model. Both judges consistently identify hallucinations and contradictions, which are the most salient failure modes for downstream user trust.
This cross-judge validation supports the coherence of clean chains used in our study.

\subsubsection{Miscalibration Rates by Condition}
\label{app:miscalibration_rates}
We report per-condition trust miscalibration rates, defined as cases where participants fail to detect a reasoning error but nevertheless report high reliance
(Agreement $\ge 5$) or high trust (Trust $\ge 5$).

\paragraph{Definition.}
We define trust miscalibration as cases in which participants fail to detect a reasoning error, yet still report high reliance on the model.
Concretely, a response is counted as miscalibrated if the participant answers \emph{No} to the error detection question while reporting either high agreement with the model’s judgment (Agreement $\ge 5$) or high self-reported trust (Trust $\ge 5$).

\paragraph{Estimation.}
Rates are computed per error type and verdict correctness condition.
We report Wilson score 95\% confidence intervals to account for binomial uncertainty.

\begin{table}[h]
\centering
\small
\caption{Miscalibration rates (missed error + high reliance/trust) by condition.
Values show proportion with 95\% CI in brackets.}
\label{tab:miscalibration_rates}
\begin{adjustbox}{width=\linewidth}
\begin{tabular}{lcc}
\toprule
\textbf{Error Type} & \textbf{Correct Verdict} & \textbf{Incorrect Verdict} \\
\midrule
\Omit{} & 0.42 [0.36, 0.48] & 0.35 [0.29, 0.41] \\
\Contr{} & 0.11 [0.07, 0.16] & 0.14 [0.09, 0.19] \\
\Hall{} & 0.09 [0.05, 0.14] & 0.12 [0.08, 0.18] \\
\bottomrule
\end{tabular}
\end{adjustbox}
\end{table}

\paragraph{Takeaway.}
Omission errors exhibit substantially higher miscalibration rates than contradictions or hallucinations across verdict conditions, confirming that omissions are both less detectable and more likely to sustain overtrust.

\subsubsection{Mixed-Effects Statistical Models and Robustness Checks}
\label{app:stats}
\paragraph{Model specifications.}
We fit mixed-effects models with scenario-level random intercepts to control for item-specific difficulty. For \emph{error detection}, which is a binary outcome, we fit a logistic mixed-effects regression model with fixed effects for verdict correctness, error type, tone, and all interactions among them, and a scenario-level random intercept to account for item-specific variability. For \emph{agreement} and \emph{trust}, which are measured on 1--7 Likert scales, we fit linear mixed-effects regression models with the same fixed-effect structure and scenario-level random intercepts.

\paragraph{Inference and multiple comparisons.}
We report fixed-effect coefficients with 95\% confidence intervals. 
Post-hoc contrasts are Holm-corrected to control for multiple comparisons.
Unless otherwise stated, reported effects are statistically significant
after correction.

\paragraph{Robustness checks.}
We verify robustness under: (i) alternative Likert binarizations (Agreement/Trust $\ge4$ vs.\ $\ge5$),
(ii) exclusion of borderline omissions, and (iii) models with additional covariates (scenario ambiguity and visual salience). Across model specifications, omissions and confident tone are the strongest predictors of trust miscalibration.

\end{document}